\documentclass[journal]{IEEEtran}

\usepackage{xcolor}
\usepackage[pdftex]{graphicx}
\usepackage{tikz}
\usetikzlibrary{decorations.pathreplacing}
\usepackage{amsmath}
\usepackage{amssymb}
\usepackage{array}
 \usepackage{url}
\usepackage{algorithmic}
\usepackage{algorithm}
\usetikzlibrary{spy,arrows}
\usepackage{hyperref}
\usepackage{xfrac}

\graphicspath{{figures/}}

\newcommand{\todo}[1]{\textcolor[rgb]{0,0,0}{#1}}
\newcommand{\tho}[1]{\textcolor[rgb]{0,0,0}{#1}}

\begin{document}
    \title{Hyperspectral and multispectral image fusion under spectrally varying spatial blurs -- Application to high dimensional infrared astronomical imaging}
  \author{Claire Guilloteau, Thomas Oberlin, Olivier Bern\'{e} and Nicolas Dobigeon
  \thanks{C. Guilloteau and N. Dobigeon are with University of Toulouse, IRIT/INP-ENSEEIHT, CNRS, 2 rue Charles Camichel, BP 7122, 31071 Toulouse Cedex 7, France (e-mail: \{Claire.Guilloteau, Nicolas.Dobigeon\}@enseeiht.fr).}
  \thanks{N. Dobigeon is also with Institut Universitaire de France (IUF).}
  \thanks{Th. Oberlin is with ISAE-SUPAERO, University of Toulouse, France (e-mail:
    Thomas.Oberlin@isae-supaero.fr).}
    \thanks{C. Guilloteau and O. Bern\'e are with Institut de Recherche en Astrophysique et Plan\'etologie (IRAP), University of Toulouse, France (e-mail:
    \{Claire.Guilloteau,Olivier.Berne\}@irap.omp.eu).}
\thanks{Part of this work has been supported by the ANR-3IA Artificial and 
Natural Intelligence Toulouse Institute (ANITI), the French
Programme Physique et Chimie du Milieu Interstellaire (PCMI) funded by the
Conseil  National  de  la  Recherche  Scientifique  (CNRS)  and  Centre  National
d'\'Etudes Spatiales (CNES).}}


\maketitle

\begin{abstract}
Hyperspectral imaging has become a significant source of valuable data for astronomers over the past decades. 
Current instrumental and observing time constraints allow direct acquisition of multispectral images, with high spatial but low spectral resolution, and hyperspectral images, with low spatial but high spectral resolution. 
To enhance scientific interpretation of the data, we propose a data fusion method which combines the benefits of each image to recover a high spatio-spectral resolution datacube. The proposed inverse problem accounts for the specificities of astronomical instruments, such as spectrally variant blurs. We provide a fast implementation by solving the problem in the frequency domain and in a low-dimensional subspace to efficiently handle the convolution operators as well as the high dimensionality of the data.
 We conduct experiments on a realistic synthetic dataset of simulated observation of the upcoming James Webb Space Telescope, and we show that our fusion algorithm outperforms state-of-the-art methods commonly used in remote sensing for Earth observation.
\end{abstract}


\section{Introduction}\label{sect:into}


\IEEEPARstart{T}{he} idea of combining spectroscopy and imaging has become very popular in the two past decades, leading to a new sensing paradigm referred to as \textit{hyperspectral} or \textit{spectral} imaging. Hyperspectral images can be thought as a whole cube of data which provides a full description of the acquired scene or sample both in space and wavelength, thus being suitable for numerous chemical or physical analyses in various applicative domains.
Hyperspectral imaging finds applications in many different fields, including remote sensing for Earth observation \cite{Chang2003book,Chang2007book} or planetology \cite{Doute2007}, material science \cite{Colliex1994,Dobigeon2012ultra,Monier2018}, dermatology \cite{DeBeule2007} and food quality monitoring \cite{Gowen2007}. In this work, we will focus on astronomy in the visible and near-infrared range. Sensing the universe in this spectral range at high spatial and spectral resolution is indeed of particular interest to study key mechanisms in astrophysics and cosmology. More specifically, this concerns for instance the combined sensing of the morphology or spectral signatures of protoplanetary disks, the interstellar medium or galaxies in the near or distant universe. For those purposes, numerous astronomical instruments have, in the past couple decades, adopted observing modes or designs allowing to acquire hyperspectral datasets. A full review of these instruments is out of the scope of this paper, but, for instance, this concerns the instruments aboard a number of space missions such as ESA's Infrared Space Observatory, NASA's Spitzer Space Telescope, or the ESA's Herschel Space Observatory, and the upcoming James Webb Space Telescope.   

However, instrumental constraints usually do not enable a direct acquisition of data-cubes combining full spatial \emph{and} spectral resolutions simultaneously (when this is the case, it is at the price of much longer integration times). A common alternative for astronomers consists in  acquiring two images of the same scene with complementary information, namely an hyperspectral (HS) image with high spectral resolution and a multispectral (MS) image with high spatial resolution. 
The HS and MS data fusion aims at combining these complementary observations to reconstruct a full data-cube at high spectral and spatial resolutions. This virtually allows to combine the performances of the data-sets at the post-processing step, without any modification of observing modes, instrumental designs, or integration times. From the astronomical point of view, these augmented data-set allow, combined to modelling, to recover detailed physical information from the scenes such as high angular resolution maps of the gas temperature and density, radiation field, metalicity, chemical abundances, etc. 


\tho{This} data fusion problem has been extensively studied in the literature of Earth observation \cite{Loncan2015, Yokoya2017}. The first methods addressed the so called pansharpening problems, which consists in fusing a MS or HS image with a panchromatic (PAN) image, i.e., a grayscale image with a single spectral band. These heuristic approaches \cite{Vivone2015, Gillespie1987} consisted in injecting spatial details extracted from the high spatial resolution image into an interpolated version of the low spatial resolution image. Those methods, in addition to be fast and easy to implement, are likely to recover spatial details with high accuracy, but they often produce significant spectral deformations~\cite{Vivone2015}.
Another class of data fusion methods is based on spectral unmixing and matrix factorization paradigms. One of the first methods was proposed in \cite{Berne2010} for fusing infrared astronomical data. According to low-rank assumption on the spectral information contained in the HS image, the latter is decomposed into two factors, representing source spectra and spatial coefficients, following a non-negative matrix factorization (NMF) \cite{Lee1999}. The source spectra matrix is then combined with a high spatial resolution coefficient matrix extracted with a non-negative least square algorithm from the MS image. The same idea has been pursued by Yokoya \emph{et al.} for remote sensing images \cite{Yokoya2012}. The so-called coupled-NMF (CNMF) method performs NMF alternatively on the HS and MS images to extract a high resolution source spectra matrix and a high resolution spatial coefficient matrix. The two methods assume linear spectral degradation and spectrally invariant spatial blur for the observations, that can be either known or estimated beforehand. The main drawback of these spectral unmixing-based fusion methods lies on their slow convergence to a local minimum, making the solution highly dependent on the initialization.
More recently, capitalizing on the prior knowledge regarding the observation instruments, the data fusion task has been formulated as an inverse problem derived from explicit forward models and complemented by appropriate spatial and/or spectral regularizations. More precisely, the forward models rely on a spectral degradation operator associated with the MS filters and a spectrally invariant spatial blurring induced by the HS sensor. Most of these methods assume a low-rank structure for the spectral information provided by the HS image. They mainly differ by the adopted spatial regularization designed to promote particular behaviors of the spatial content. For instance, a convex regularization as a form of vector total variation has been used in \cite{Simoes2015}, promoting sparsity in the distribution of the gradient of the reconstructed image. Therefore, this fused image is expected to be spatially smooth, except for a small number of areas, coinciding with sharp edges. Instead of promoting a smooth content, the regularization introduced in \cite{Wei2015_2}, represents the target image as a sparse combination of elements of a dictionary composed of spatial patches and learned from the MS image. The resulting optimization problems are solved iteratively thanks to particular instances of the alternating direction  method of multipliers \cite{Afonso2010}. More recently, the authors in \cite{Wei2015, Wei2016} show that such fusion inverse problems can be formulated as a Sylvester equation and solved analytically, significantly decreasing the computational complexity of the aforementioned iterative methods.

However, all these techniques are not suitable to tackle the fusion of high dimensional astronomical data. The first challenge is to handle the high dimensionality of the data, considerably larger than the usual dimension encountered in remote sensing. Indeed, a high spatio-spectral fused image in Earth remote sensing is composed of at most a few hundreds of spectral bands while spatio-spectral astronomical data are typically composed of up to several thousands, or even tens of thousands of spectral measurements. 
Moreover, the spatial resolution of space- or airborne Earth observations is mainly limited by atmosphere turbulence \cite{Pearson1976}. Nevertheless, the spatial resolution of spaceborne astronomical observations is limited by diffraction. This limit is wavelength dependent and can be estimated by the Rayleigh criterion \cite{Rayleigh1880}. It defines the angular resolution $\theta = 1.220 \frac{\lambda}{D}$, where $\lambda$ is the wavelength of the light and $D$ the diameter of the aperture. In practice, this physical property means that the operators associated with spatial blurs should be considered as spectrally varying while restoring astronomical MS and HS images \cite{Soulez2013,HadjYoucef2017}. This crucial issue significantly increases the complexity of the forward models and make the fusion methods previously discussed inoperative. Indeed, as mentioned above, the forward models commonly used for Earth observation data fusion rely on a spectrally invariant spatial blur to describe the HS observation and a subsampling operator combined with a spectral degradation operator for the MS observation. The main contributions reported in this work tackle both challenges: we design a fusion method and its fast implementation suitable for fusing large-scale astronomical data while taking into account the specificities of astronomical imaging, in particular the spectrally variant blur underlying the MS and HS observations.

The paper is organized as follows. Section~\ref{sect:pb_form} describes the observational forward models and introduces the fusion inverse problem. Then, Section~\ref{sect:fast_imp} presents our main contribution: a fast implementation to solve the inverse problem. To this end, the optimization problem is rewritten in the frequency domain, while an appropriate vectorization step enables to formulate the spatial degradations in a low-dimensional subspace. In Section~\ref{sect:results}, the performance of the proposed method is assessed using a realistic simulated astrophysical dataset and compared qualitatively and quantitatively with state of the art methods. Section \ref{sect:conclusion} finally concludes the paper.

\section{Problem formulation}\label{sect:pb_form}

\subsection{Forward models}\label{sect:fwd_models}
This section derives the mathematical models associated with two observation instruments providing images of complementary spatial and spectral resolutions. The first one is an optical imager which acquires a MS image of high spatial resolution denoted $\mathbf{Y}_\mathrm{m} \in \mathbb{R}^{l_\mathrm{m}\times p_\mathrm{m}}$, where $l_\mathrm{m}$ and $p_\mathrm{m}$ denote the numbers of spectral bands and pixels, respectively.
The second instrument is a spectrometer which acquires a full HS data-cube $\mathbf{Y}_\mathrm{h} \in \mathbb{R}^{l_\mathrm{h}\times p_\mathrm{h}}$ of lower spatial resolution, with $l_\mathrm{m}<l_\mathrm{h}$ and $p_\mathrm{h}<p_\mathrm{m}$. From these measurements, the objective of the fusion process is to recover a HS image of high spatial resolution denoted $\mathbf{X} \in \mathbb{R}^{l_\mathrm{h} \times p_\mathrm{m}}$, which has the same spatial resolution as the MS image and the same spectral resolution of the HS one.
The responses of the two sensors are modeled by a series of linear transformations that describe successive spatial and spectral degradations of light emerging from the scene of interest. With the adopted ordering of the elements in the matrix $\mathbf{X}$, spectral and spatial degradations will be represented as left and right operators, respectively. More precisely, we assume that the MS and HS images result from the following forward models
\begin{equation}
    \mathbf{Y}_\mathrm{m} \approx \mathbf{L}_\mathrm{m} \mathcal{M}(\mathbf{X})
    \label{eq:ym}
\end{equation}
\begin{equation}
    \mathbf{Y}_\textrm{h} \approx \mathbf{L}_\textrm{h}\mathcal{H}(\mathbf{X})\mathbf{S} 
    \label{eq:yh}
\end{equation}
where the  symbol $\approx$ accounts for random noises and model mismodeling, and the other operators are detailed hereafter.
First, $\mathbf{L}_\mathrm{m} \in \mathbb{R}^{l_\mathrm{m}\times l_\mathrm{h}}$ and $\mathbf{L}_\mathrm{h} \in \mathbb{R}^{l_\mathrm{h}\times l_\mathrm{h}}$ are spectral degradation operators, respectively associated with MS and HS images. The MS observation instrument integrates the spectral bands of the initial scene $\mathbf{X}$ over the spectral dimension to provide each MS band. The lines of the matrix $\mathbf{L}_\mathrm{m}$ in \eqref{eq:ym} are thus made of the transmission functions of the $l_\mathrm{m}$ corresponding filters. On the other hand, the spectral information of the initial scene $\mathbf{X}$ is attenuated by the optical system of the HS instrument. Therefore, $\mathbf{L}_\mathrm{h}$ is a diagonal matrix made of the spectral transmission function of the instrument. 
Second, $\mathcal{M} : \mathbb{R}^{l_\mathrm{h}\times p_\mathrm{m}} \to \mathbb{R}^{l_\mathrm{h}\times p_\mathrm{m}}$ in \eqref{eq:ym} and $\mathcal{H} : \mathbb{R}^{l_\mathrm{h}\times p_\mathrm{m}} \to \mathbb{R}^{l_\mathrm{h}\times p_\mathrm{m}}$ in \eqref{eq:yh} are spatial degradation operators which model the blurs caused by the optical system of both instruments. In the context of astronomical imaging addressed in this work, we can reasonably assume the associated point spread functions (PSFs) to be space-invariant, but they strongly depend on the wavelength, following a Rayleigh criterion \cite{Rayleigh1880}. Therefore, $\mathcal{M}(\cdot)$ and $\mathcal{H}(\cdot)$ are 2D spatial convolution operators with spectrally variant blurring kernels specific to each instrument. 
Finally, the spatial resolution of the HS image is impaired by a subsampling operator $\mathbf{S} \in \mathbb{R}^{p_\mathrm{m}\times p_\mathrm{h}}$ with an integer decimation factor $d$ such that $p_\mathrm{h} = \frac{p_\mathrm{m}}{d^2}$.  In other words, right-multiplying by $\mathbf{S}$ amounts to keeping one pixel over $d^2$.
In this work, we assume that all the operators are known.

\subsection{Inverse problem}

To recover $\mathbf{X}$ from the two noisy observations, we adopt the general framework of (variational) inverse problem, trying to fit the observations while adding regularization terms to promote prior knowledge on the sought solution. By denoting $(\cdot)^H$ the Hermitian transpose and $\|\cdot\|_{\mathrm{F}}^2 = \mathrm{Tr}\left((\cdot)(\cdot)^H\right)$ the Frobenius norm,
this amounts to solving the generic problem
\begin{multline}
\hat{\mathbf{X}} = \underset{\mathbf{X}}{\text{argmin}} \left( \frac{1}{2\sigma_\mathrm{m}^2} \|\mathbf{Y}_\textrm{m} -\mathbf{L}_\textrm{m}\mathcal{M}(\mathbf{X})\|_\mathrm{F}^2 \right. \\ 
\left. + \frac{1}{2\sigma_\mathrm{h}^2} \|\mathbf{Y}_\textrm{h} -\mathbf{L}_\textrm{h}\mathcal{H}(\mathbf{X})\mathbf{S}\|_\mathrm{F}^2 + \varphi_\mathrm{spec}(\mathbf{X}) + \varphi_\mathrm{spac}(\mathbf{X}) \right)
\label{eq:problem1}
\end{multline}
where the two first terms are data fidelity terms related respectively to the MS and the HS images. Minimizing these data fidelity terms is equivalent to maximize the log-likelihood associated to a white Gaussian noise model in the data, i.e., the symbols $\approx$ in \eqref{eq:ym} and \eqref{eq:yh} stand for additive corruptions $\mathbf{N}_\textrm{m}$ and $\mathbf{N}_\textrm{h}$ assumed to be independent white Gaussian noise with variance $\sigma_\textrm{m}^2$ and $\sigma_\textrm{h}^2$, respectively. Although this hypothesis may be not realistic for astronomical images as they are known to be rather corrupted by a mixed Poisson-Gaussian noise \cite{Starck2006}, the least-square loss is chosen for the sake of computational efficiency. It is worth noting that the experimental results reported in Section \ref{sect:results} will show that this simplifying assumption does not significantly impair the relevance of the proposed method.

Besides, the terms $\varphi_\mathrm{spec}(\cdot)$ and $\varphi_\mathrm{spac}(\cdot)$ in \eqref{eq:problem1} stand for spectral and spatial regularizations, respectively. Regarding $\varphi_\mathrm{spec}(\cdot)$, HS image bands are known to be highly correlated. Thus the pixels of the full scene $\mathbf{X}$ can be reasonably assumed to live in a subspace whose dimension $l_\textrm{sub}$ is much smaller than its spectral dimension $l_\textrm{h}$. This property can be formulated by imposing a low-rank structure on the scene $\mathbf{X}$ to be recovered, i.e., $\mathbf{X}=\mathbf{V}\mathbf{Z}$  where the columns of $\mathbf{V} \in \mathbb{R}^{l_\mathrm{h}\times l_\mathrm{sub}}$ (with $l_\mathrm{sub} \leq l_\mathrm{h}$) spans the signal subspace and $\mathbf{Z} \in \mathbb{R}^{l_\mathrm{sub}\times p_\mathrm{m}}$ gathers the corresponding representation coefficients. This decomposition implicitly imposes a spectral regularization, since the spectra of the fused image are assumed to be linear combinations of the reference spectra defining $\mathbf{V}$. The subspace of interest spanned by $\mathbf{V}$ can be fixed beforehand thanks to prior knowledge regarding the scene of interest or estimated from the HS measurements, e.g., after conducting a principal component analysis (PCA). A similar strategy has been widely adopted in numerous works of the literature dedicated to hyperspectral image enhancement \cite{Wycoff2013,Simoes2015,Wei2015}. Another asset of this change of variable lies in a significant reduction of the complexity of the optimization problem since \emph{i)} estimating the decomposition coefficients $\hat{\mathbf{Z}}$ is sufficient to recover the fused image $\hat{\mathbf{X}}=\mathbf{V}\hat{\mathbf{Z}}$ and \emph{ii)} this decomposition allows the forward models to be rewritten in the subspace spanned by $\mathbf{V}$, which leads to a scalable algorithm (see Section \ref{sect:vecto} for details).

Concerning the spatial regularization term $\varphi_\mathrm{spac}(\cdot)$, it is based on the assumption that the sought image is \emph{a priori} spatially smooth, in agreement with typical scenes encountered in astrophysical observations. We thus propose to minimize the energy of the spatial discrete gradient of the image, also known as Sobolev regularization. This writes
\begin{equation*}
    \varphi_\mathrm{spac}(\mathbf{Z}) =\mu\|\mathbf{Z}\mathbf{D}\|_\mathrm{F}^2
\end{equation*}
where the matrix $\mathbf{D}$ stands for a $1$st order 2-D finite difference operator and the regularization parameter $\mu \ge 0$ controls the strength of the regularization.
The problem now becomes
\begin{multline}
\hat{\mathbf{Z}} = \underset{\mathbf{Z}}{\text{argmin}} \left( \frac{1}{2\sigma_\mathrm{m}^2} \|\mathbf{Y}_\textrm{m} -\mathbf{L}_\textrm{m}\mathcal{M}(\mathbf{VZ})\|_\mathrm{F}^2 \right. \\ 
\left. + \frac{1}{2\sigma_\mathrm{h}^2} \|\mathbf{Y}_\textrm{h} -\mathbf{L}_\textrm{h}\mathcal{H}(\mathbf{VZ})\mathbf{S}\|_\mathrm{F}^2 + \mu\|\mathbf{Z}\mathbf{D}\|_\mathrm{F}^2 \right).
\label{eq:pinv}
\end{multline} 
The next section presents an efficient algorithmic scheme designed to solve the minimization problem \eqref{eq:pinv}.

\section{Fast implementation}\label{sect:fast_imp}

Although quadratic, the problem stated in \eqref{eq:pinv} cannot be easily solved by conventional methods such as fast gradient descent~\cite{Beck2009} or conjugate gradient \cite{Shewchuk1994} because of the spectrally variant blurs in $\mathcal{M(\cdot)}$ and $\mathcal{H(\cdot)}$. Indeed, resorting to such algorithms needs to evaluate the gradient at each iteration, requiring the application of operators $\mathcal{M}(\cdot)$ and $\mathcal{H}(\cdot)$ and their respective adjoints, i.e., applying a set of $4l_\mathrm{h}$ distinct 2D spatial convolutions. Storing and processing thousands of distinct PSFs would  annihilate the benefit of the dimension reduction induced by the low-rank decomposition underlying by the matrix $\mathbf{V}$.
In this section, we propose a fast implementation tailored to this fusion task under spectrally variant blurring. We first fully formulate the problem in the frequency domain to handle the heavy convolution operators $\mathcal{M}$ ans $\mathcal{H}$. Subsequently, we vectorize and combine these operators to solve the problem in the low-dimensional subspace spanned by the columns of $\mathbf{V}$. Ultimately, we provide and discuss the computational complexity of the proposed implementation, highlighting the benefits of these contributions.

\subsection{Resolution in the frequency domain}\label{sect:freq}

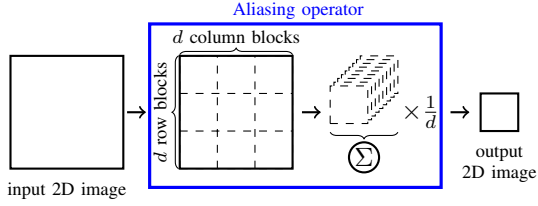
\begin{figure}
    \centering
    \begin{tikzpicture}[decoration={brace,raise=-1}, scale=0.5]
    
    \draw[thick] (-0.5,0) rectangle ++(3,3);
    \node[] at (1,-0.60) {\scriptsize \begin{tabular}{c} input 2D image \end{tabular}};
    
    \draw[dashed] (4,0) grid (7,3);
    \draw[thick] (4,0) rectangle (7,3);
    \draw[decorate] (4,3.1) -- (7,3.1) node[above,pos=0.5] {\scriptsize $d$ column blocks};
    \draw[decorate,decoration={brace,raise=-1,mirror}] (3.9,3) -- (3.9,0) node[above,pos=0.5,rotate=90] {\scriptsize $d$ row blocks};
    
    \draw[dashed,fill=white] (8.8,2) rectangle (9.8,3);
    \draw[dashed,fill=white] (8.7,1.9) rectangle (9.7,2.9);
    \draw[dashed,fill=white] (8.6,1.8) rectangle (9.6,2.8);
    \draw[dashed,fill=white] (8.5,1.7) rectangle (9.5,2.7);
    \draw[dashed,fill=white] (8.4,1.6) rectangle (9.4,2.6);
    \draw[dashed,fill=white] (8.3,1.5) rectangle (9.3,2.5);
    \draw[dashed,fill=white] (8.2,1.4) rectangle (9.2,2.4);
    \draw[dashed,fill=white] (8.1,1.3) rectangle (9.1,2.3);
    \draw[dashed,fill=white] (8,1.2) rectangle (9,2.2);
    \draw[thick] (8.9,0.3) circle (0.4);
    \draw[decorate,decoration={brace,raise=-1,mirror}] (8,0.9) -- (9.8,0.9);
    \node[] at (8.9,0.3) {$\Sigma$};
    \node[] at (10.4,1.5) {$\times \frac{1}{d}$};
    
    \draw[thick] (12,1) rectangle ++(1,1);
    \node[] at (12.5,0.15) {\scriptsize \begin{tabular}{c} output \\ 2D image  \end{tabular}};
    
    \draw[very thick, color=blue] (3.2,-0.5) rectangle ++(7.75,4.35);
    \node[color=blue, very thick] at (7.15,4.2) {\scriptsize Aliasing operator};
    
    \draw[->,thick] (2.6,1.5) -- ++(0.5,0);
    \draw[->,thick] (7.25,1.5) -- ++(0.5,0);
    \draw[->,thick] (11.2,1.5) -- ++(0.5,0);
    
    \end{tikzpicture}
    \caption{Illustration of the aliasing operation $\mathbf{\dot{S}}$.}
    \label{fig:aliasing}
\end{figure}

It is widely admitted that computing convolutions in the frequency domain using fast Fourier transform (FFT) and its inverse (iFFT) can be faster than directly convolving in the spatial domain. Here, we propose to reformulate the problem in the Fourier domain to benefit from this computational advantage. Indeed, every spatial degradation operator (convolution, subsampling and finite differences operators) can be expressed or approximated in the Fourier domain by diagonal operators (i.e., pointwise multiplications), thus reducing the  computation burden. First, under periodic boundary assumptions, the set of 2D spatial convolutions in $\mathcal{H}(\cdot)$ and $\mathcal{M}(\cdot)$ can be achieved by cyclic convolutions acting on $\mathbf{VZ}$. We denote by $\odot$ the element-wise matrix multiplication and $\mathbf{F}$ the 2D-discrete Fourier transform (DFT) matrix ($\mathbf{F}\mathbf{F}^H=\mathbf{F}^H\mathbf{F}=\mathbf{I}$) such that $\mathbf{\dot{Z}}=\mathbf{Z}\mathbf{F}$. Thus, the convolution at a specific spectral band $l$ can be rewritten 
\begin{equation*}
    [\mathcal{M}(\mathbf{VZ})]^l = \left(\mathbf{\dot{M}}^l\odot [\mathbf{V\dot{Z}}]^l \right) \mathbf{F}^H
\end{equation*}
\begin{equation*}
    [\mathcal{H}(\mathbf{VZ})]^l = \left(\mathbf{\dot{H}}^l\odot [\mathbf{V\dot{Z}}]^l \right) \mathbf{F}^H
\end{equation*}
where $\mathbf{\dot{M}}^l$ and $\mathbf{\dot{H}}^l$ denote the 2D-DFTs of the $l\mathrm{th}$ PSFs related to, respectively, the multi- and the hyperspectral observation instrument.
The down-sampling operator $\mathbf{S}$ of factor $d$  can be written in the Fourier domain as an aliasing operator $\mathbf{\dot{S}} = \mathbf{S}\mathbf{F}$ of factor $d$, which sums $d^2$ blocks of an input matrix, as illustrated in Fig.~\ref{fig:aliasing}. Similarly to the downsampling operator, the aliasing operator acts independently on every spectral band. Each 2D spatial image with $p_\mathrm{m}$ pixels is partitioned into $d^2$ blocks and these blocks are summed up to produce a 2D spatial image with $p_\mathrm{h} = \frac{p_\mathrm{m}}{d^2}$ pixels. 

Regarding the spatial regularization, the $1$st order 2D finite differences operator $\mathbf{D}$ can be seen as a 2D convolution operator with kernels $\begin{pmatrix} 1 & -1 \end{pmatrix}$ and $\begin{pmatrix} 1 \\ -1 \end{pmatrix}$. This operator needs to be applied to the low-dimensional representation maps $\mathbf{Z}$, whose spectral dimension $l_{\mathrm{sub}}$ is much smaller than $\mathbf{V}\mathbf{Z}$. Thus, the computational gain reached by computing this regularization in the Fourier domain remains negligible. However, for practical reasons and to simplify the implementation, we decide to adopt this strategy. More precisely, again, under cyclic boundary conditions, this regularization term can be expressed in the Fourier domain as a term-wise multiplication such that $$\mathbf{ZD} = \left( \mathbf{\dot{Z}} \odot \mathbf{\dot{D}} \right) \mathbf{F}^H.$$

Finally, following Parseval's identity, the problem \eqref{eq:pinv} is fully rewritten in the Fourier domain
\begin{multline}
\widehat{\mathbf{\dot{Z}}} = \underset{\mathbf{\dot{Z}}}{\text{argmin}} \left(\frac{1}{2\sigma_\mathrm{m}^2}\left\|\mathbf{\dot{Y}}_\mathrm{m} -\mathbf{L}_\mathrm{m}((\mathbf{V}\mathbf{\dot{Z}})\odot\mathbf{\dot{M}})\right\|_\mathrm{F}^2 \right. \\
\left. + \frac{1}{2\sigma_\mathrm{h}^2} \left\|\mathbf{\dot{Y}}_\mathrm{h} - \mathbf{L}_\mathrm{h}((\mathbf{V}\mathbf{\dot{Z}})\odot\mathbf{\dot{H}})\mathbf{\dot{S}}\right\|_\mathrm{F}^2+ \mu \left\| \mathbf{\dot{Z}} \odot \mathbf{\dot{D}}\right\|_\mathrm{F}^2\right)
\label{eq:pinv2}
\end{multline}
where $\mathbf{\dot{Y}}_\mathrm{m} = \mathbf{Y}_\mathrm{m} \mathbf{F}$ and $\mathbf{\dot{Y}}_\mathrm{h} = \mathbf{Y}_\mathrm{h} \mathbf{F}$. Finally the fused image can be obtained as $\hat{\mathbf{X}} = \mathbf{V}\widehat{\mathbf{\dot{Z}}}\mathbf{F}^H$.

\subsection{Vectorization}\label{sect:vecto}

The second step consists in computing the sequence of operators in the subspace spanned by the columns of $\mathbf{V}$ instead of being applied to the full image $\mathbf{V\dot{Z}}$. To do so, we introduce the lexicographically ordered counterparts $\mathbf{\dot{y}}_\mathrm{m}$, $\mathbf{\dot{y}}_\mathrm{h}$, $\underline{\mathbf{V}}$ and $\mathbf{\dot{z}}$ of $\mathbf{\dot{Y}}_\mathrm{m}$, $\mathbf{\dot{Y}}_\mathrm{h}$, $\mathbf{V}$ and $\mathbf{\dot{Z}}$, respectively, such that
\begin{align*}
\mathbf{\dot{y}}_\mathrm{m} &= \left[\mathbf{\dot{Y}}_\mathrm{m}^1, \cdots, \mathbf{\dot{Y}}_\mathrm{m}^{l_\mathrm{m}}\right]^T 
&\mathbf{\dot{y}}_\mathrm{h} &=\left[\mathbf{\dot{Y}}_\mathrm{h}^1, \cdots, \mathbf{\dot{Y}}_\mathrm{h}^{l_\mathrm{h}}\right]^T \\
\mathbf{\underline{V}} &= \mathbf{V} \otimes  \mathbf{I}_{p_\mathrm{m}} &\mathbf{\dot{z}} &= \left[\mathbf{\dot{Z}}^1, \cdots, \mathbf{\dot{Z}}^{l_\mathrm{sub}}\right]^T
\end{align*}
where $\otimes$ denotes the Kronecker product and $\mathbf{I}_{p}$ is the $p\times p$ identity matrix.   
With these notations, the problem~ \eqref{eq:pinv2} is equivalent to 
\begin{multline*}
\widehat{\mathbf{\dot{z}}}=\underset{\mathbf{\dot{z}}}{\text{argmin}} \left(\frac{1}{2\sigma_\mathrm{m}^2}\|\mathbf{\dot{y }}_\mathrm{m} -\underline{\mathbf{L}}_\mathrm{m}\mathbf{\underline{\dot{M}}}\underline{\mathbf{V}}\mathbf{\dot{z}}\|_2^2 \right. \\
\left. + \frac{1}{2\sigma_\mathrm{h}^2} \|\mathbf{\dot{y }}_\mathrm{h} - \underline{\mathbf{\dot{S}}}\underline{\mathbf{L}}_\mathrm{h}\underline{\mathbf{\dot{H}}}\mathbf{\underline{V}}\mathbf{\dot{z}}\|_2^2+ \mu \|\underline{\mathbf{\dot{D}}}\mathbf{\dot{z}}\|_2^2\right)
\end{multline*}
where $\underline{\mathbf{L}}_\mathrm{m}$, $\underline{\mathbf{L}}_\mathrm{h}$, $\underline{\mathbf{\dot{M}}}$, $\underline{\mathbf{\dot{H}}}$, $\underline{\mathbf{\dot{S}}}$ and $\underline{\mathbf{\dot{D}}}$ are vectorized forms of $\mathbf{L}_\mathrm{m}$, $\mathbf{L}_\mathrm{h}$, $\mathbf{\dot{M}}$, $\mathbf{\dot{H}}$, $\mathbf{\dot{S}}$ and $\mathbf{\dot{D}}$, respectively, such that
\begin{equation*}
    \underline{\mathbf{L}}_\mathrm{m} \underline{\mathbf{\dot{M}}} \mathbf{\underline{V}} \mathbf{\dot{z}} = \left(
    \begin{array}{c}
        \left[\mathbf{L}_\mathrm{m}((\mathbf{V}\mathbf{\dot{Z}})\odot\mathbf{\dot{M}})\right]^1 \\
        \vdots  \\
        \left[\mathbf{L}_\mathrm{m}((\mathbf{V}\mathbf{\dot{Z}})\odot\mathbf{\dot{M}})\right]^{l_\mathrm{m}}
    \end{array}
    \right) 
\end{equation*}
and
\begin{equation*}
    \underline{\mathbf{\dot{S}}}\underline{\mathbf{L}}_\mathrm{h} \underline{\mathbf{\dot{H}}} \mathbf{\underline{V}} \mathbf{\dot{z}} = \left(
    \begin{array}{c}
         \left[\mathbf{L}_\mathrm{h}((\mathbf{V}\mathbf{\dot{Z}})\odot\mathbf{\dot{H}})\mathbf{\dot{S}}\right]^1  \\
    \vdots \\
    \left[\mathbf{L}_\mathrm{h}((\mathbf{V}\mathbf{\dot{Z}})\odot\mathbf{\dot{H}})\mathbf{\dot{S}}\right]^{l_\mathrm{h}}
    \end{array}
    \right).
\end{equation*}
The structures and expressions of all vectorized spatial and spectral operators are detailed in Appendix~\ref{app:vectorization}. 
Finally, the fusion task boils down to solving the linear system
\begin{equation}
    \mathbf{A} \mathbf{\dot{z}} = \mathbf{b}
   \label{eq:pinv3}
\end{equation}
where $\mathbf{A} \in \mathbb{R}^{l_\mathrm{sub}p_\mathrm{m} \times l_\mathrm{sub}p_\mathrm{m}}$ and $\mathbf{b} \in \mathbb{R}^{l_\mathrm{sub}p_\mathrm{m}}$ are defined by
\begin{align}
\label{eq:amatrix}
    \nonumber\mathbf{A}&= \frac{1}{\sigma_\mathrm{m}^2}\mathbf{\underline{V}}^H\underline{\mathbf{\dot{M}}}^H\underline{\mathbf{L}}_\mathrm{m}^H\underline{\mathbf{L}}_\mathrm{m}\underline{\mathbf{\dot{M}}}\mathbf{\underline{V}} \\
    &+ \frac{1}{\sigma_\mathrm{h}^2}\mathbf{\underline{V}}^H\underline{\mathbf{\dot{H}}}^H\underline{\mathbf{L}}_\mathrm{h}^H\underline{\mathbf{\dot{S}}}^H \underline{\mathbf{\dot{S}}}\underline{\mathbf{L}}_\mathrm{h}\underline{\mathbf{\dot{H}}}\mathbf{\underline{V}} + \mu \underline{\mathbf{\dot{D}}}^H\underline{\mathbf{\dot{D}}},\\
\label{eq:bmatrix}
    \mathbf{b}&= - \frac{1}{\sigma_\mathrm{m}^2}\mathbf{\underline{V}}^H\underline{\mathbf{\dot{M}}}^H\underline{\mathbf{L}}_\mathrm{m}^H\mathbf{\dot{y }}_\mathrm{m}
    - \frac{1}{\sigma_\mathrm{h}^2}\mathbf{\underline{V}}^H\underline{\mathbf{\dot{H}}}^H\underline{\mathbf{L}}_\mathrm{h}^H\underline{\mathbf{\dot{S}}}^H\mathbf{\dot{y }}_\mathrm{h}.
    \end{align}
Interestingly, as suggested by \eqref{eq:pinv3}~and explicitly expressed by \eqref{eq:amatrix} and \eqref{eq:bmatrix}, the quantities $\mathbf{A}$ and $\mathbf{b}$ resort to all spatial and spectral operators. In particular, they combine the individual wavelength-dependent PSFs defining $\mathcal{H}(\cdot)$ and $\mathcal{M}(\cdot)$ to be jointly expressed in the low-dimensional subspace through the left-composition by the projection operator $\underline{\mathbf{V}}^H$. Moreover, the symmetric matrix $\mathbf{A}$ is sparse and composed of, at most, $d^2 l_\mathrm{sub}^2 p_\mathrm{m}$ non-zero entries, i.e., only a $\sfrac{d^2}{p_\mathrm{m}}$-th proportion of the matrix coefficients is non-zero, arranged according to a very particular structure detailed in Appendix~\ref{app:matrix_A}. As a consequence, its high level of sparsity, combined with its block structure, allows the matrix $\mathbf{A}$ to be computed only once as a pre-processing step and cheaply stored in memory (see Appendix~\ref{app:matrix_A} for a detailed description of its computation). Finally, this matrix can be easily called out along the iterations of a gradient-based descent algorithm implemented to solve \eqref{eq:pinv3}. It is also worth noting that this matrix only depends on the forward models defined by the observation instruments, the adopted spatial regularization and the matrix $\mathbf{V}$ spanning the signal subspace. Thus, once this subspace does not change, this matrix does not need to be recomputed to fuse multiple sets of MS and HS measurements. 

\subsection{Complexity analysis}

This section discusses the complexity imposed by one iteration for three different gradient descent algorithms that solve the fusion problem. More precisely, we compare a naive implementation minimizing \eqref{eq:pinv}, the so-called frequency algorithm minimizing the problem \eqref{eq:pinv2} formulated in the Fourier domain and the proposed algorithm solving the vectorized formulation yielding the linear system \eqref{eq:pinv3}. The respective complexities are expressed as functions of the spatial and spectral dimensions of the data to be fused, namely $p_\mathrm{m}$, $l_\mathrm{h}$ and $l_\mathrm{m}$, and the intrinsic dimension $l_{\mathrm{sub}}$ of the subspace. They are reported in Table~\ref{tab:complexity} for the general case, i.e., without assuming any particular prevalence of one of these quantities over the others. However, for typical scenarios arising in the applicative context of astronomical imaging that will be considered in the experiments (see Section \ref{sect:results}), we have $l_{\mathrm{sub}} \leq l_{\mathrm{m}} \leq \log p_{\mathrm{m}}$. In this context, the following findings can be drawn.  


When considering a naive implementation, the heaviest computational burden to solve \eqref{eq:pinv} directly results from evaluating the gradient of the corresponding quadratic cost function, which amounts to $\mathcal{O}(l_\mathrm{h}p_\mathrm{m} \log p_\mathrm{m})$ operations. Note that this implementation relies on cyclic convolutions operated in the Fourier domain but requires back and forth in the image domain by FFT and inverse FFT at each iteration. When the problem is fully formulated in the Fourier domain (see Section \ref{sect:freq}), the cost of computing the gradient associated to \eqref{eq:pinv2} reduces to $\mathcal{O}(l_\mathrm{h}p_\mathrm{m} l_\mathrm{m})$. By vectorizing the problem (see Section \ref{sect:vecto}), the gradient is directly given by the matrix $\mathbf{A}$ in \eqref{eq:amatrix}. Thus the core steps of the iterative algorithm solving \eqref{eq:pinv3} consist in matrix-vector products, which requires only $\mathcal{O}(p_\mathrm{m}l_{\mathrm{sub}}^2)$ operations thanks to the high level of sparsity of $\mathbf{A}$. Consequently, one iteration of this vectorized implementation is significantly less complex than the naive and Fourier domain-based resolutions.

Besides, while the naive implementation does not require any pre-processing step, the two alternative schemes proposed in Sections \ref{sect:freq} and \ref{sect:vecto} rely on quantities computed beforehand. More precisely, to solve \eqref{eq:pinv2} in the frequency domain, FFT of the MS and HS images and PSFs are required, for a overall complexity of $\mathcal{O}(l_\mathrm{h} p_\mathrm{m} \log p_\mathrm{m})$. In addition, solving \eqref{eq:pinv3} requires to compute the matrix $\mathbf{A}$ in \eqref{eq:amatrix} and the vector $\mathbf{b}$ in  \eqref{eq:bmatrix}. Specifically, the most heavy step is computing the two first terms in the right-hand side of \eqref{eq:amatrix}, for a overall complexity of $\mathcal{O}(l_\mathrm{h}p_\mathrm{m} l_\mathrm{sub}^2)$. 
Therefore, the pre-processing step involved in the vectorized implementation is more time-consuming but this step is performed only once before solving the problem iteratively. Moreover, as already highlighted, this pre-processing is significantly lightened when fusing several sets of HS and MS measurements since only $\mathbf{b}$ needs to be updated, provided the spatial regularization and the signal subspace remain unchanged.

\begin{table}[]
\renewcommand{\arraystretch}{1.5}
    \centering
    \caption{Asymptotic complexity (as $\mathcal{O}(\cdot)$): one iteration of the gradient-based algorithm and pre-processing.} 
    \begin{tabular}{|c|cc|}
    \hline
    Method & Iteration & Pre-processing \\
    \hline
    \hline
    Naive & $l_\mathrm{h}p_\mathrm{m}  \max\left\{l_{\mathrm{sub}}, l_\mathrm{m}, \log p_\mathrm{m}\right\}$ & -- \\
    \hline  
    Frequency & $l_\mathrm{h}p_\mathrm{m} \max\left\{l_{\mathrm{sub}}, l_\mathrm{m}\right\}$  & $l_\mathrm{h}p_\mathrm{m} \log p_\mathrm{m}$\\
    \hline
    Vectorized & $p_\mathrm{m}l_\mathrm{sub}^2$ & $l_\mathrm{h}p_\mathrm{m} l_\mathrm{sub}^2$ \\
    \hline
    \end{tabular}
    \vspace{0.5cm}
    \label{tab:complexity}
\end{table}

Beyond the computational complexity, given the high dimensionality of the problem, issues raised by handling the data and instrument models should be also discussed. In particular, loading the entire fused product and all spectrally variant PSFs  is impossible in high dimension when using conventional computing ressources. As a consequence, when solving the fusion problem with the naive strategy or in the frequency domain (see Section \ref{sect:freq}), computing $(\mathbf{V}\mathbf{\dot{Z}})\odot\mathbf{\dot{M}}$ and $(\mathbf{V}\mathbf{\dot{Z}})\odot\mathbf{\dot{H}}$ at each iteration of the descent algorithm requires on-the-fly loading of each PSF, which increases  the computational times significantly. Conversely, the vectorized implementation requires to load these PSFs only once during the pre-processing.


\section{Experimental results}\label{sect:results}

This section assesses the performance of the proposed fusion method when applied to a simulated yet realistic  astronomical dataset. This dataset is discussed in the next paragraph. The considered figures-of-merit, compared methods and quantitative and qualitative results are reported subsequently. 

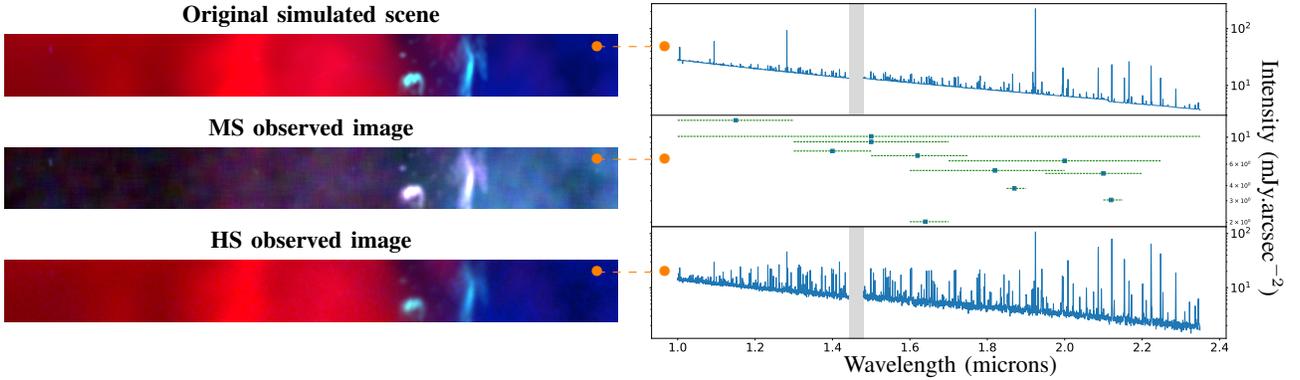
\begin{figure*}
    \centering
    \begin{tikzpicture}
    \node[inner sep=0pt] (graph) at (0,0) {\scalebox{1}[-1]{\includegraphics[width=0.45\linewidth]{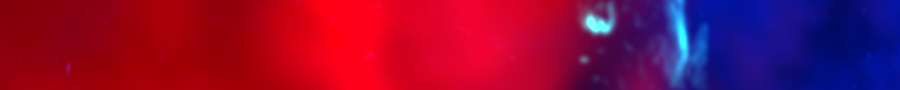}}};
    \node[] at (0, 0.65) {\small \bf Original simulated scene};
    \node[inner sep=0pt] (graph) at (0,-1.5) {\scalebox{1}[-1]{\includegraphics[width=0.45\linewidth]{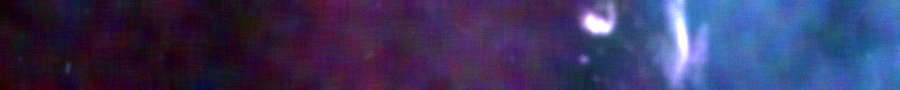}}};
    \node[] at (0, -0.85) {\small \bf MS observed image};
    \node[inner sep=0pt] (graph) at (0,-3) {\scalebox{1}[-1]{\includegraphics[width=0.45\linewidth]{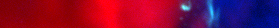}}};
    \node[] at (0, -2.35) {\small \bf HS observed image};
    
    \node[inner sep=0pt] (graph) at (8.5,-1.5)
        {\includegraphics[width=0.45\linewidth]{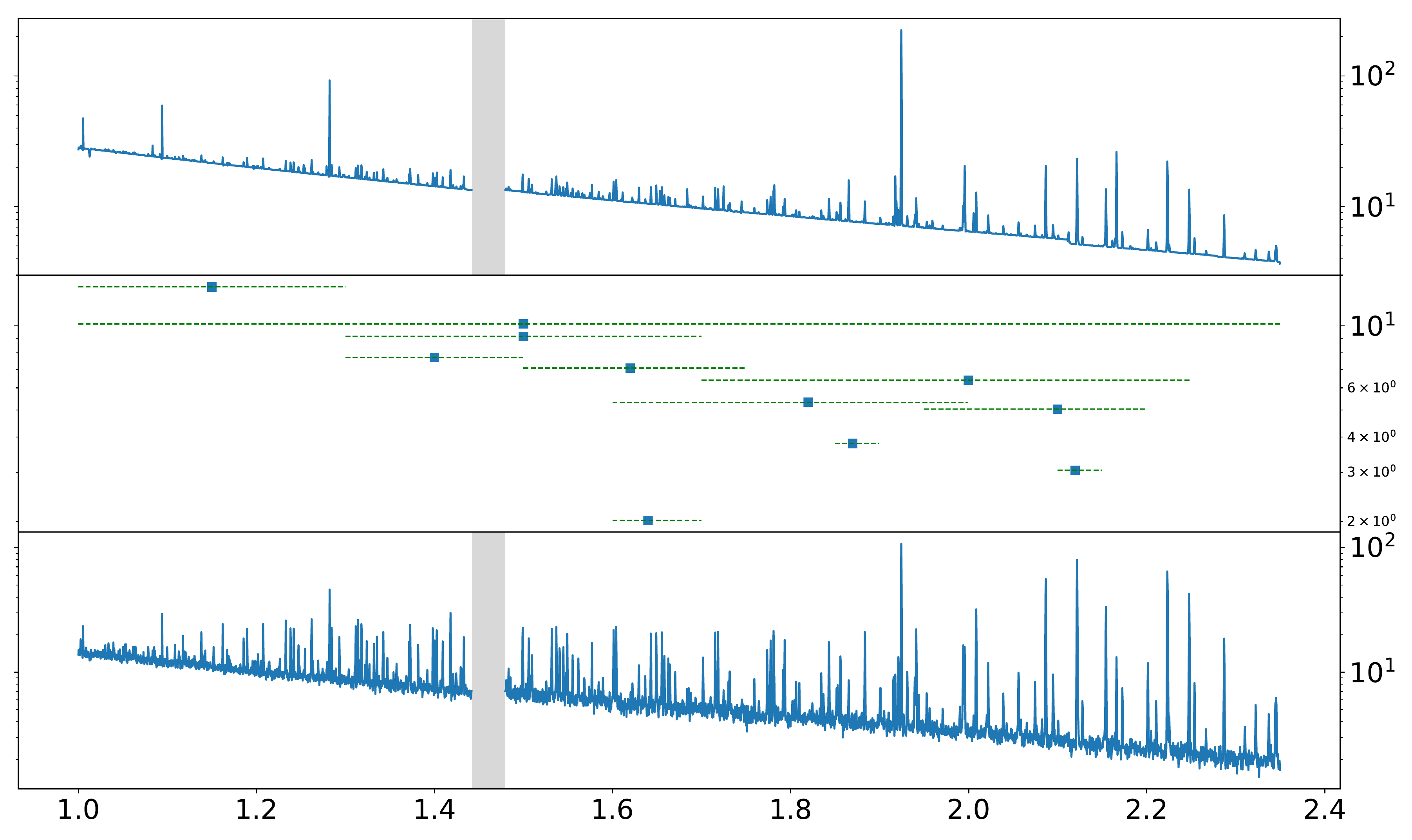}};
    \node[] at (8.5, -4) {\small Wavelength (microns)};
    \node[rotate=-90] at (12.75, -1.5) {\small Intensity (mJy.arcsec$^{-2}$)};
        
    \draw[orange, dashed] (3.8,0.25) -- ++(0.9,0);
    \node[orange] at (3.8,0.25) {$\bullet$};
    \node[orange] at (3.8+0.9,0.25) {$\bullet$};
    
    \draw[orange, dashed] (3.8,0.25-1.5) -- ++(0.9,0);
    \node[orange] at (3.8,0.25-1.5) {$\bullet$};
    \node[orange] at (3.8+0.9,0.25-1.5) {$\bullet$};
    
    \draw[orange, dashed] (3.8,0.25-3) -- ++(0.9,0);
    \node[orange] at (3.8,0.25-3) {$\bullet$};
    \node[orange] at (3.8+0.9,0.25-3) {$\bullet$};
    
    \end{tikzpicture}
    \caption{Left: RGB compositions of the synthetic simulated scene (top), the NIRCam Imager MS image (middle) and the NIRSpec IFU HS image (bottom) [Red channel: H$_2$ emission line pic intensity at $2.122\mu$m, Green channel: H recombination line pic intensity at $1.865\mu$m, Blue channel: Fe$^+$ emission line pic intensity at $1.644\mu$m]. Right:  A spectrum from 1.0 to 2.35 microns related to a pixel of each image on their left. From top to bottom, the first two are original spectra from the synthetic scene with 4974 points, the following two are observed spectra from the multiband image provided by the NIRCam Imager forward model with 11 spectral points, the last two are calibrated observed spectra from the HS image provided by the NIRSpec IFU forward model with about 5000 spectral points.}
    \label{fig:orig_hs_ms}
\end{figure*}

\subsection {Simulated dataset}

The simulated dataset considered in the experiments was specifically designed to assess multi- and hyperspectral data fusion in the particular context of high dimensional astronomical observations performed by the James Webb Space Telescope (JWST). The generation process is accurately described in \cite{Guilloteau2019_2} and more briefly recalled hereafter. This dataset is composed of a high spatial and high spectral resolution synthetic scene of a photodissociation region (PDR) located in the Orion Bar. This scene is accompanied with a pair of corresponding simulated MS and HS observations. The resolution of the synthetic scene matches the spectral resolution of the HS instrument and the spatial resolution of the MS sensor, and its field of view and spectral range corresponds to plausible real acquisitions that will be performed by the JWST.

The synthetic scene has been generated under a low-rank assumption such that its constitutive spectra are linear mixtures of 4 synthetic elementary spectra spatially distributed according to 4 maps representing the spatial abundances of each elementary spectrum over the scene. To simulate the expected spatial and spectral content of the Orion bar, four real images acquired by different telescopes are combined to build the spatial maps and the spectral signatures of the elementary components were chosen to be those likely present in this region (see \cite{Guilloteau2019_2} for more details). This simulated scene will be denoted  $\mathbf{X}$ in the following and will represent the reference (i.e., ground-truth) data-cube we aim to recover by fusing the HS and MS measurements. It is composed of $90 \times 900$ pixels and $4974$ spectral bands ranging from 1 to 2.35 $\mu$m. 

\begin{figure}[h!]
    \centering
    \begin{tikzpicture}
    \begin{scope}
        \clip(-7, 0) rectangle ++(3.5,3.5);
        \node[inner sep=0pt] (graph) at (-0.5,0)
        {\includegraphics[width=1.5\linewidth]{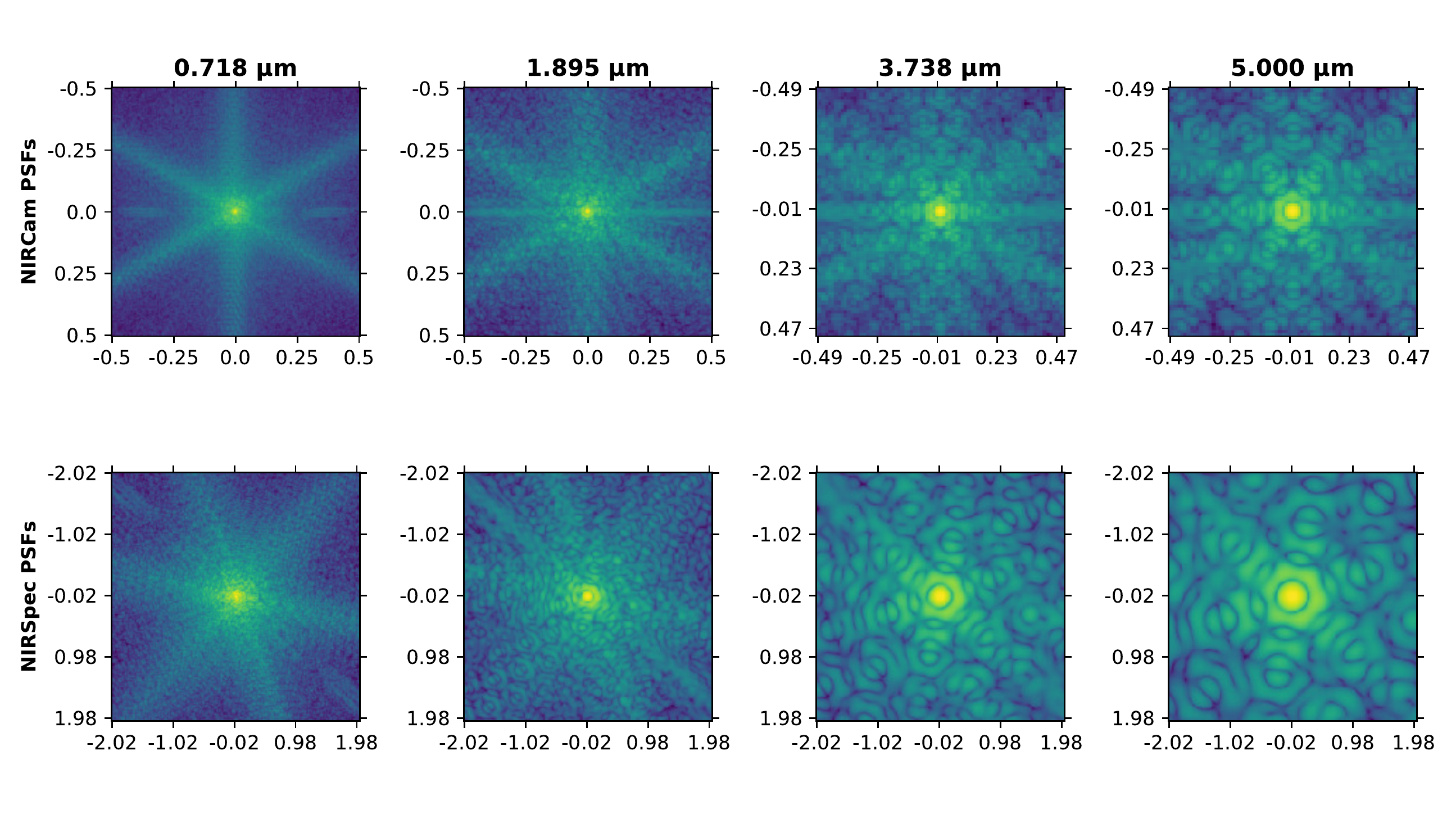}};
    \end{scope}
    \begin{scope}
        \clip(-3, 0) rectangle ++(3.5,3.5);
        \node[inner sep=0pt] (graph) at (-6.5,0)
        {\includegraphics[width=1.5\linewidth]{psfs-eps-converted-to.pdf}};
    \end{scope}
    \begin{scope}
        \clip(-7, -4) rectangle ++(3.5,3.5);
        \node[inner sep=0pt] (graph) at (-0.5,0)
        {\includegraphics[width=1.5\linewidth]{psfs-eps-converted-to.pdf}};
    \end{scope}
    \begin{scope}
        \clip(-3, -4) rectangle ++(3.5,3.5);
        \node[inner sep=0pt] (graph) at (-6.5,0)
        {\includegraphics[width=1.5\linewidth]{psfs-eps-converted-to.pdf}};
    \end{scope}
    \node[] at (-5,0.1) {\small Offset (arsec)};
    \node[] at (-5,-3.5) {\small Offset (arsec)};
    \node[] at (-1.45,0.1) {\small Offset (arsec)};
    \node[] at (-1.45,-3.5) {\small Offset (arsec)};
    \node[inner sep=0pt] (graph) at (0.5,1.8)
    {\includegraphics[width=0.07\linewidth]{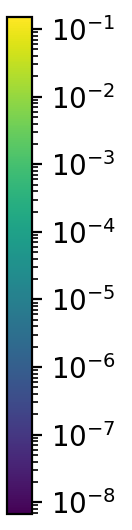}};
    \node[inner sep=0pt] (graph) at (0.5,-1.7)
    {\includegraphics[width=0.065\linewidth]{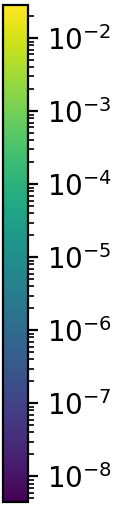}};
    \end{tikzpicture}
    \caption{PSFs of the NIRCam Imager (top) and NIRSpec IFU (bottom) calculated with \emph{webbpsf} \cite{Perrin2012} for two particular wavelengths (logarithmic scale).}
    \label{fig:psfs}
\end{figure}

The corresponding MS and HS observed images were simulated from this reference synthetic image following the forward models introduced in Section~\ref{sect:fwd_models}, where the spatial and spectral degradation operators are those of the JWST instrumentation documentation\footnote{Instrumental documentation available on STScI website: \url{https://jwst-docs.stsci.edu/}}.
The MS image $\mathbf{Y}_\mathrm{m}$ simulates the output of the near-infrared camera (NIRCam) imager and is composed of $90 \times 900$ pixels and $11$ spectral bands. The HS image $\mathbf{Y}_\mathrm{h}$ consists of  $30 \times 300$ pixels and $4974$ spectral bands with the specificities of the integral field unit (IFU) of the near-infrared spectrograph  (NIRSpec). The spatial subsampling factor $d$ is thus set to $d=3$.
The spectral degradation operators $\mathbf{L}_\mathrm{m}$ and $\mathbf{L}_\mathrm{h}$ are the spectral responses of those two instruments as specified by the documentation. 
The 2-D spatial convolution operators $\mathcal{M}(\cdot)$ and $\mathcal{H}(\cdot)$ are each composed of 4974 PSFs whose full width at half-maximum (FWHM) is linearly varying with wavelength. Therefore, the widest PSF is 2.35 times larger than the thinnest. For multi- and hyperspectral observation instruments, they are of size $161\times 161$ pixels and $145 \times 145$ pixels, respectively and, because of the specific shape of JWST mirrors, these PSFs are strongly anisotropic, as illustrated in Fig.~\ref{fig:psfs}. This figure emphasizes the crucial need of accounting for spectrally variant spatial convolution operators in the two forward models. 

Finally, the simulated images $\mathbf{Y}_\mathrm{m}$ and $\mathbf{Y}_\mathrm{h}$ include a realistic Poisson-Gaussian mixed noise which is expected to corrupt astronomical data. They are first corrupted with a Poisson noise approximated by a multiplicative Gaussian noise of mean and variance the photon count in each pixel. The instrumental so-called readout noise is subsquently modeled by an additive spatially correlated Gaussian noise, with mean and covariance matrix depending on instruments and readout patterns, assumed to be known. 

Red-green-blue (RGB) color compositions (left) and spectra (right) of the reference synthetic image (top), the simulated MS observed image (middle) and simulated HS observed image (bottom) are shown in Fig.~\ref{fig:orig_hs_ms}. Each color in the composite images is associated to a specific emission line chemically related to a particular region of the PDR to highlight the various structures of the scene. Spectra in the right-hand side of the figure coincide with a pixel in the dark-blue region. Those illustrations show how the signal is degraded by the instruments. For the MS observations, the RGB composition shows less contrast, due to the loss of spectral
information induced by the filters. On the other hand, the hyperspectral data is clearly less spatially resolved, and the spectrum exhibits a lower signal-to-noise ratio (SNR).

\subsection {Quality metrics}

The performances of the compared data fusion algorithms are assessed according to three reconstruction quality measures. We propose to evaluate the spectral distortion between reconstructed and target spectra through the average spectral angle mapper (SAM), defined by  
$$\text{aSAM}(\mathbf{\hat{X}},\mathbf{X})= \frac{1}{l_\mathrm{h}} \sum_{p=1}^{l_\mathrm{h}} \text{arccos} \left(\frac{\langle\mathbf{X}_p,\mathbf{\hat{X}}_p\rangle}{\|\mathbf{X}_p\|_2\|\mathbf{\hat{X}}_p\|_2}\right)$$ where $\mathbf{\hat{X}}_p$ is a reconstructed spectrum and $\mathbf{X}_p$ is the corresponding reference spectrum.
The structural similarity (SSIM) index is then used to estimate the degradation of spatial structural information. The SSIM index is defined as  
$$\text{SSIM}(\mathbf{\hat{X}}^l,\mathbf{X}^l)=   \frac{\left(2\mu_{\mathbf{\hat{X}}^l}\mu_{\mathbf{X}^l} + C_1\right)\left(2\sigma_{\mathbf{\hat{X}^l}\mathbf{X}^l}+C_2\right)}{\left(\mu_{\mathbf{\hat{X}}^l}^2 + \mu_{\mathbf{X}^l}^2 + C_1\right)\left(\sigma_{\mathbf{\hat{X}^l}}^2 + \sigma_{\mathbf{X^l}}^2 +C_2\right)}$$
where $\mathbf{\hat{X}}^l$ is a $l$th reconstructed spectral band, $\mathbf{X}^l$ is the corresponding reference spectral band and  $\mu_{\mathbf{\hat{X}}^l}$,  $\mu_{\mathbf{X}^l}$, $\sigma_\mathbf{\hat{X}^l}^2$, $\sigma_{\mathbf{X^l}}^2$, $\sigma_{\mathbf{\hat{X}^l}\mathbf{X}^l}$ are empirical statistics defined in \cite{Wang2004} and $C_j \propto L^2$ ($j=1,2$) is the dynamic range of $\mathbf{X}^l$. In this paper, we rather consider the average complementary SSIM (acSSIM) across all bands defined by $$ \text{acSSIM} (\mathbf{\hat{X}},\mathbf{X}) = 1 - \frac{1}{l_{\mathrm{m}}}\sum_{l=1}^{l_{\mathrm{m}}}\text{SSIM} (\mathbf{\hat{X}}^l,\mathbf{X}^l). $$
Finally, the overall peak SNR (PSNR) measures the overall reconstruction quality in the least-square sense: 
$$ \text{PSNR}(\mathbf{\hat{X}},\mathbf{X})= 10\log_{10}\left(\frac{\max 
    (\mathbf{X})}{\|\mathbf{X}-\mathbf{\hat{X}}\|_\mathrm{F}^2}\right) $$
where $\mathbf{\hat{X}}$ is the reconstructed image and $\mathbf{X}$ is the reference.
Note that a good performance is achieved when both the aSAM and acSSIM are low while the PSNR is large. All these quantities have been averaged over 20 Monte-Carlo runs.

\subsection {Compared methods}


We first consider a naive super-resolution method relying on a low-rank assumption, referred to afterwards as the baseline method. This approach consists in spatially upsampling the projection of the HS image onto the subspace spanned by the columns of $\mathbf{V}$ with a bi-cubic spline interpolation to reach the spatial resolution of the MS image. 

We also compare our fusion algorithm, designated as ``Proposed'', with two methods widely known for fusing MS and HS or MS and PAN remote sensing data: the Brovey method \cite{TeMing2001} and the robust fast fusion using a Sylvester equation (R-FUSE) \cite{Wei2016}.
The first one is a component substitution approach originally designed to fuse MS and PAN images. It interpolates the projection of the HS image over the spectral subspace to the spatial resolution of the MS image and injects extracted details from the MS image. It only requires the prior knowledge of the spectral blur operator $\mathbf{L}_\mathrm{m}$. The second method formulates the fusion task as an inverse problem derived from forward models of observation instruments complemented with a Gaussian prior. This problem uses a spectral degradation operator $\mathbf{L}_\mathrm{m}$ related to the multispectral instrument and a spectrally invariant PSF related to the hyperspectral instrument. \todo{In these experiments, this unique PSF is chosen as the PSF corresponding to the mean-energy wavelength.} The problem is written as a Sylvester equation and solved analytically, substantially decreasing the computational complexity. 

Finally to evaluate the relevance of the fusion task, we also compare our fusion algorithm with its two non-symmetric versions, where one of the data-fit term is removed. The first version, called MS-only, solves the following spectral deconvolution problem 
$$ \hat{\mathbf{Z}} = \underset{\mathbf{Z}}{\text{argmin}} \left(\frac{1}{2\sigma_\mathrm{m}^2} \|\mathbf{Y}_\mathrm{m} - \mathbf{L}_\mathrm{m}\mathcal{M}(\mathbf{V}\mathbf{Z})\|_\mathrm{F}^2+ \mu_\mathrm{m} \| \mathbf{Z} \mathbf{D}\|_\mathrm{F}^2\right) $$
where only the data fitting term related to the MS image is considered.
\todo{Similarly, the second version, called HS-only and similar to \cite{HadjYoucef2018}, solves the following HS super-resolution problem including only the data fitting term related to the HS image}
$$ \hat{\mathbf{Z}} = \underset{\mathbf{Z}}{\text{argmin}} \left(\frac{1}{2\sigma_\mathrm{h}^2} \|\mathbf{Y}_\mathrm{h} - \mathbf{L}_\mathrm{h}\mathcal{H}(\mathbf{V}\mathbf{Z})\mathbf{S}\|_\mathrm{F}^2+ \mu_\mathrm{h} \| \mathbf{Z} \mathbf{D}\|_\mathrm{F}^2\right).$$

All the aforementioned methods require a subspace identification to find the basis matrix $\mathbf{V}$. This step is performed by PCA conducted on the HS image, as it is expected to contain all the relevant spectral information. These methods also require an hyperparameter setting. In this paper, the hyperparameter is set such that it leads to the highest PSNR value and the lowest aSAM and acSSIM. 

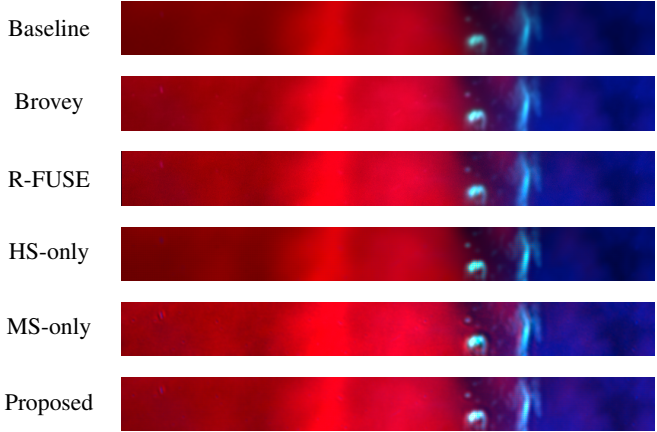
\begin{figure}[h!]
    \centering
    \begin{tikzpicture}
    \node[inner sep=0pt] (graph) at (0,0) {\scalebox{1}[-1]{\includegraphics[width=0.8\linewidth]{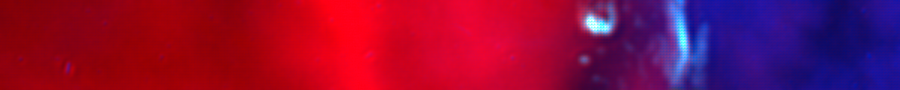}}};
    \node[inner sep=0pt] (graph) at (0,1) {\scalebox{1}[-1]{\includegraphics[width=0.8\linewidth]{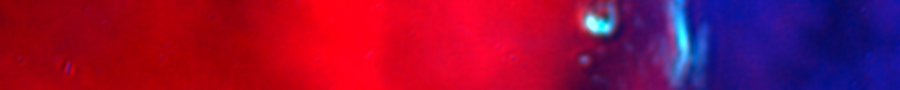}}};
    \node[inner sep=0pt] (graph) at (0,2) {\scalebox{1}[-1]{\includegraphics[width=0.8\linewidth]{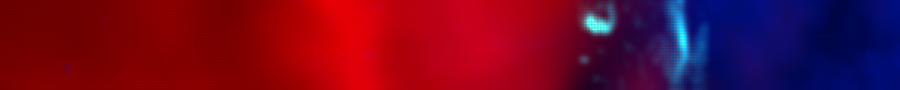}}};
    \node[inner sep=0pt] (graph) at (0,3) {\scalebox{1}[-1]{\includegraphics[width=0.8\linewidth]{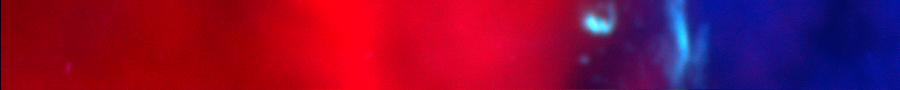}}};
    \node[inner sep=0pt] (graph) at (0,4) {\scalebox{1}[-1]{\includegraphics[width=0.8\linewidth]{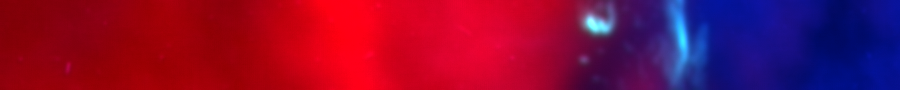}}};
    \node[inner sep=0pt] (graph) at (0,5) {\scalebox{1}[-1]{\includegraphics[width=0.8\linewidth]{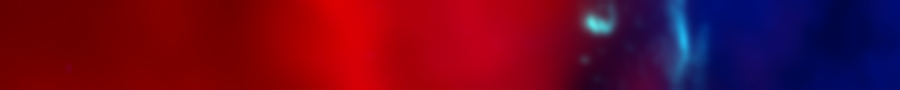}}};

    \node[] at (-4.5,5) {\small Baseline};
    \node[] at (-4.5,4) {\small Brovey};
    \node[] at (-4.5,3) {\small R-FUSE};
    \node[] at (-4.5,2) {\small HS-only};
    \node[] at (-4.5,1) {\small MS-only};
    \node[] at (-4.5,0) {\small Proposed};
    
    \end{tikzpicture}
    \caption{From top to bottom: RGB compositions of fused images reconstructed by the baseline, Brovey, R-FUSE, HS-only, MS-only and proposed method. The color composition is the same as for Fig. \ref{fig:orig_hs_ms} (left).}
    \label{fig:results}
\end{figure}

\subsection {Results}

\begin{figure}[h!]
    \centering
    \begin{tikzpicture}

    \node[inner sep=0pt] (graph) at (-3,5) {\scalebox{1}[-1]{\includegraphics[width=0.3\linewidth]{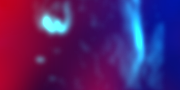}}};
    \node[inner sep=0pt] (graph) at (0,5) {\scalebox{1}[-1]{\includegraphics[width=0.3\linewidth]{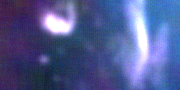}}};
     \node[inner sep=0pt] (graph) at (3,5) {\scalebox{1}[-1]{\includegraphics[width=0.3\linewidth]{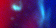}}};
    \node[inner sep=0pt] (graph) at (-3,3) {\scalebox{1}[-1]{\includegraphics[width=0.3\linewidth]{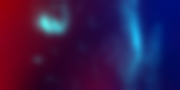}}};
    \node[inner sep=0pt] (graph) at (0,3) {\scalebox{1}[-1]{\includegraphics[width=0.3\linewidth]{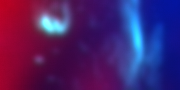}}};
    \node[inner sep=0pt] (graph) at (3,3) {\scalebox{1}[-1]{\includegraphics[width=0.3\linewidth]{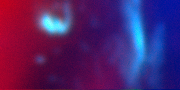}}};
    \node[inner sep=0pt] (graph) at (-3,1) {\scalebox{1}[-1]{\includegraphics[width=0.3\linewidth]{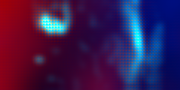}}};
    \node[inner sep=0pt] (graph) at (0,1) {\scalebox{1}[-1]{\includegraphics[width=0.3\linewidth]{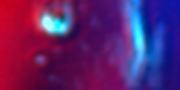}}};
    \node[inner sep=0pt] (graph) at (3,1) {\scalebox{1}[-1]{\includegraphics[width=0.3\linewidth]{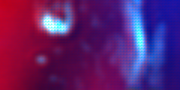}}};

    \node[] at (-3,5.9) {\small Original};
    \node[] at (0,5.9) {\small MS observed image};
    \node[] at (3,5.9) {\small HS observed image};
    \node[] at (-3,3.9) {\small Baseline};
    \node[] at (0,3.9) {\small Brovey};
    \node[] at (3,3.9) {\small R-FUSE};
    \node[] at (-3,1.9) {\small HS datafit};
    \node[] at (0,1.9) {\small MS datafit};
    \node[] at (3,1.9) {\small Proposed};
    
    \end{tikzpicture}
    \caption{Zooms on strong structures excerpt from Fig. \ref{fig:orig_hs_ms} (simulated, MS observed and HS observed images) and from Fig. \ref{fig:results} (fused images by the compared methods).}
    \label{fig:results_zoom}
\end{figure}
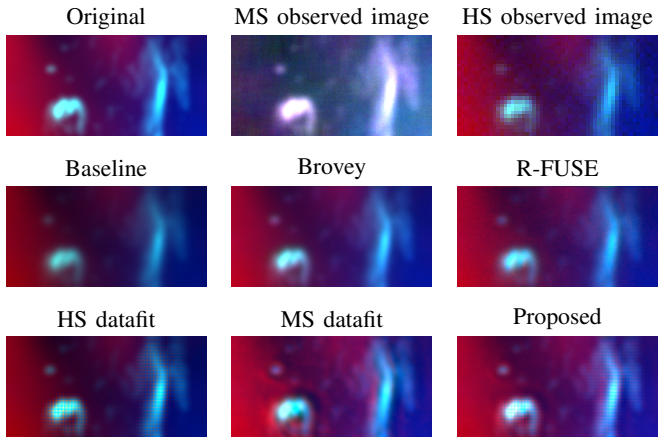

The fusion results obtained by the six compared methods are depicted in Fig.~\ref{fig:results} as RGB images using the same color composition as in Fig. \ref{fig:orig_hs_ms}. Zooms on sharp structures in the scene are shown in Fig.~\ref{fig:results_zoom}. 
Qualitatively, the reconstruction appears to be excellent.  
Denoising seems to be efficient for most methods, with a slightly noisier fused product obtained with R-FUSE. The baseline and HS-only methods are not able to restore the energy in the signal while the MS-only and proposed methods appear to better recover even very high intensities, especially on sharp edges, as shown in Fig.~\ref{fig:results_zoom}. The gain in spectral and spatial resolution of reconstructed images of the proposed algorithm with respect to MS and HS images respectively is clearly noticeable. The contrast between color components in the MS observed image is restored, as well as spatial details blurred in the HS observed image.

To better assess method performances, quantitative results are reported in Table~\ref{tab:results_highsnr}. The two best results for each measure are highlighted in bold. As expected, the HS-only method shows a very low aSAM, i.e., an excellent spectral reconstruction but a poor spatial reconstruction with the second worst cSSIM index. On the other side, the MS-only method provides the best spatial reconstruction but the worst spectral reconstruction. Our method provides, as a trade-off between HS-only and MS-only, the second best spatial and spectral reconstructions. The best overall PSNR values are reached by MS-only, our proposed method and HS-only, improving the baseline performance up to $8$dB. State-of-the-art methods give similar quantitative results, with a slightly better PSNR value for the Brovey method.
Pre-processing time aside, all the methods are very fast and perform data fusion in less than $30$ seconds.

\begin{table}[h!]
\renewcommand{\arraystretch}{1.5}
    \centering
    \caption{Performance of fusion methods: aSAM (rad), acSSIM, PSNR (dB), and Time (pre-processing + fusion, seconds).}
    \begin{tabular}{|c|cccc|}
    \hline
    Methods & aSAM & acSSIM & PSNR & Time \\
    \hline
    \hline
    Baseline & 0.0296 & 0.0428 & 66.88 & / \\
    \hline
    Brovey & 0.0304 & 0.0040 & 71.04 & 17 \\
    \hline
    R-FUSE & 0.0360 & 0.0036 & 69.68 & 26 \\
    \hline
    HS-only & \bf 0.0118 & 0.0239 & 72.90 & 1600 + 20 \\
    \hline
    MS-only & 0.0389 & \bf 0.0018 & \bf 75.00 & 600 + 15 \\
    \hline
    Proposed & \bf 0.0247 & \bf 0.0029 & \bf 74.90 & 2200 + 20 \\
    \hline
    \end{tabular}
    \vspace{0.5cm}
    \label{tab:results_highsnr}
\end{table}

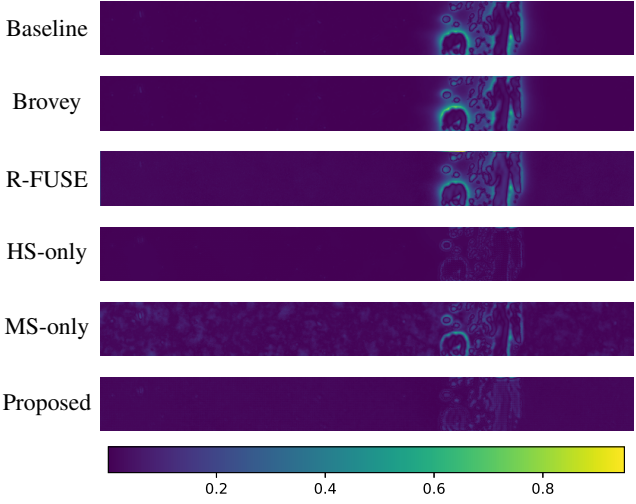
\begin{figure}[h!]
    \centering
    \begin{tikzpicture}
    \node[inner sep=0pt] (graph) at (0,0) {\includegraphics[width=0.8\linewidth]{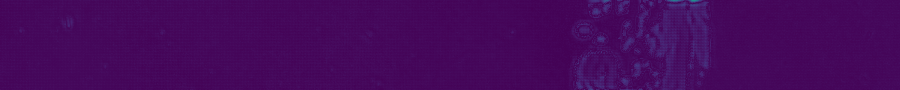}};
    \node[inner sep=0pt] (graph) at (0,1) {\includegraphics[width=0.8\linewidth]{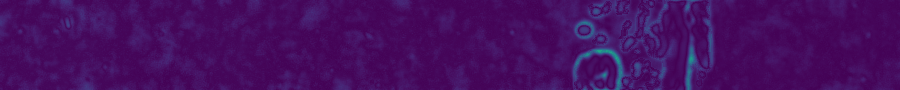}};
    \node[inner sep=0pt] (graph) at (0,2) {\includegraphics[width=0.8\linewidth]{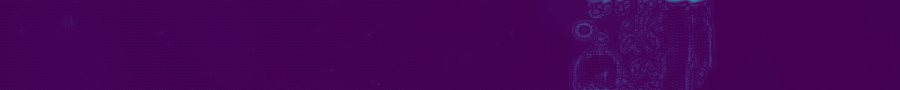}};
    \node[inner sep=0pt] (graph) at (0,3) {\includegraphics[width=0.8\linewidth]{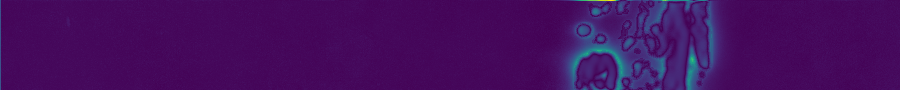}};
    \node[inner sep=0pt] (graph) at (0,4) {\includegraphics[width=0.8\linewidth]{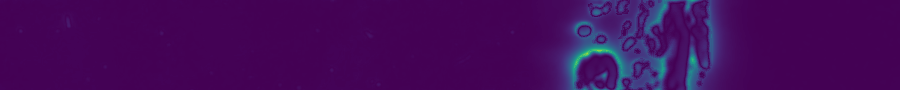}};
    \node[inner sep=0pt] (graph) at (0,5) {\includegraphics[width=0.8\linewidth]{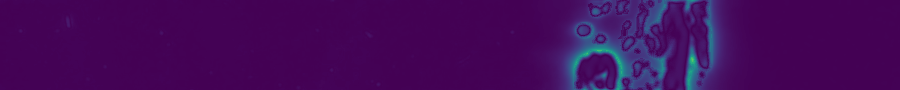}};

    \node[] at (-4.25,5) {\small Baseline};
    \node[] at (-4.25,4) {\small Brovey};
    \node[] at (-4.25,3) {\small R-FUSE};
    \node[] at (-4.25,2) {\small HS-only};
    \node[] at (-4.25,1) {\small MS-only};
    \node[] at (-4.25,0) {\small Proposed};
    
    \begin{scope}
        \clip(-4,-1.25) rectangle (4,-0.5);
        \node[inner sep=0pt] (graph) at (-0.125,1.5) {\includegraphics[width=\linewidth]{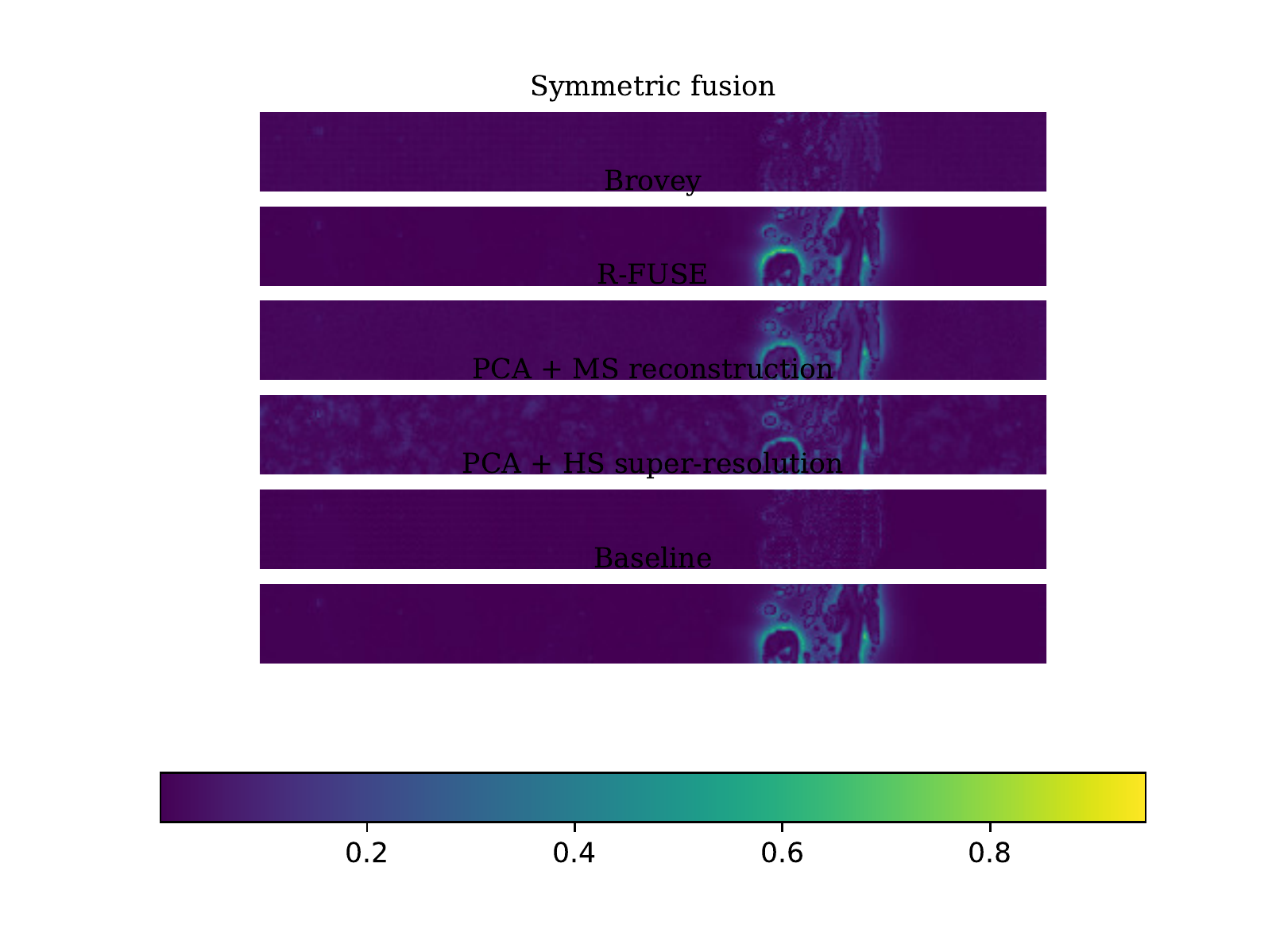}};
    \end{scope}
    \end{tikzpicture}
    \caption{Spatial maps of the SAM obtained by, from top to bottom, the baseline, Brovey, R-FUSE, HS-only, MS-only and proposed method. The smaller SAM, the better the reconstruction.}
    \label{fig:sam_maps}
\end{figure}

\begin{figure}
    \centering
    \includegraphics[width=\linewidth]{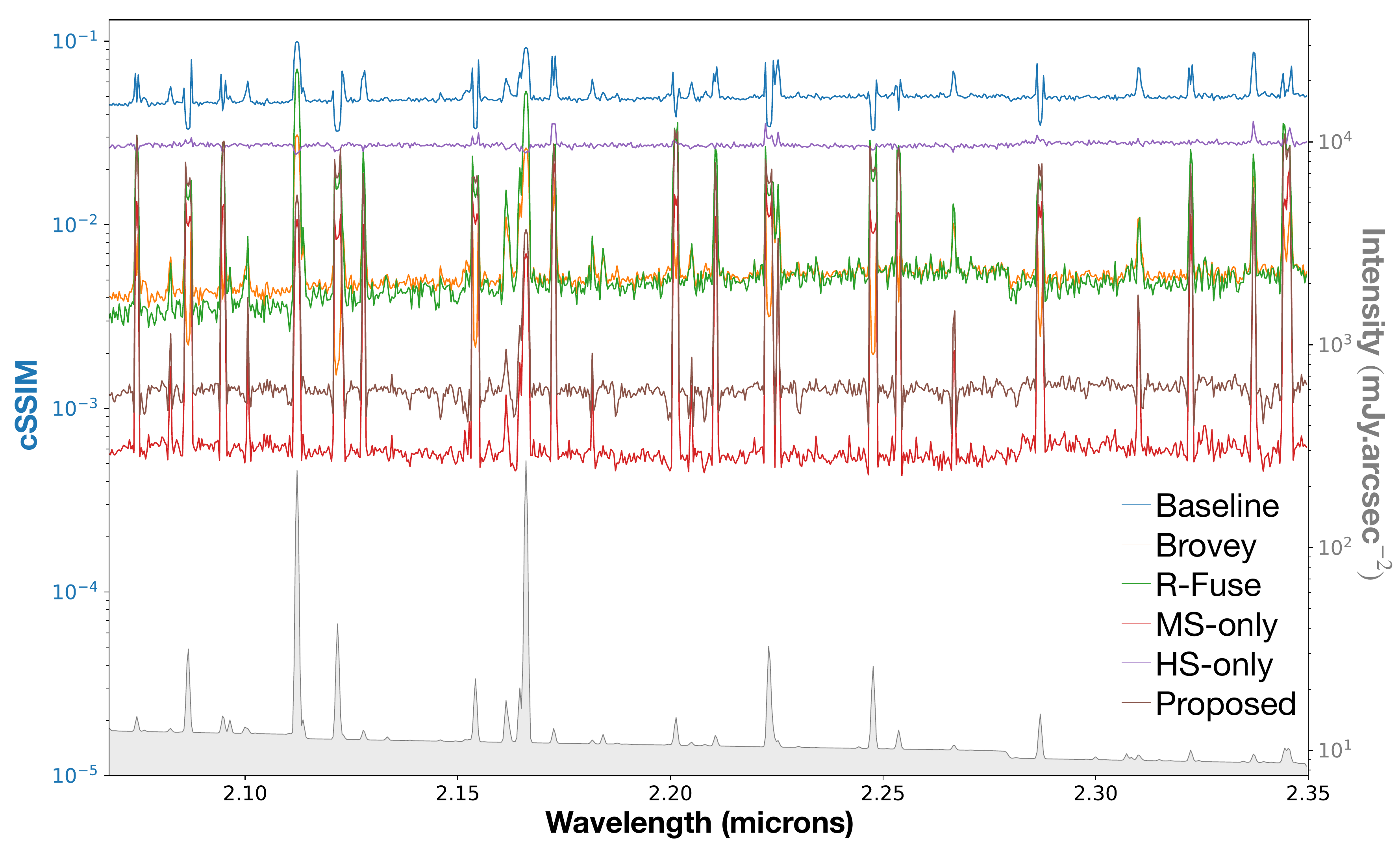}
    \caption{Color lines: cSSIM as a function of the wavelength obtainted by the compared methods. The smaller cSSIM, the better the reconstruction. Gray line and shaded area: a spectrum located in the reference scene around a sharp structure.}
    \label{fig:cssim_spec}
\end{figure}

Fig.~\ref{fig:sam_maps} presents SAM errors maps. These spectral errors have been calculated between reference and reconstructed spectra without averaging over the pixels. This figure highlights that, for each method, spectra located around a sharp structure of the scene (i.e., characterized by a high gradient region) show bad reconstructions (yellow pixels). This bad reconstruction is even worse for baseline, Brovey, R-FUSE and MS-only methods while HS-only and the proposed method provide the lowest SAM maxima. For most methods, this can be explained by the adopted regularizations, which promote spatially smooth content and therefore distribute the flux over neighboring pixels, leading to higher SAM values. On the contrary, in spatially smooth regions, all the methods present a very low spectral error.
Fig.~\ref{fig:cssim_spec} represents cSSIM errors as function of the wavelength, i.e., without averaging over the spectral bands. Baseline and HS-only methods show very large spatial errors whereas MS-only and the proposed method provide the best cSSIM values. In between, Brovey and R-FUSE present intermediate and slightly increasing with wavelength cSSIM values. This may be explained by the fact that R-FUSE exploits a unique PSF, i.e., neglecting the spectrally spatial blur affecting the data. Indeed, larger wavelength bands are blurrier than short wavelength bands and therefore spatially worse reconstructed with an inappropriate model. The reference spectrum displayed in the bottom of the graph emphasizes that variations in cSSIM w.r.t. wavelength are correlated with high intensity emission lines in the scene. This is also likely due to the regularization term which tends to favor smooth images especially for high intensity spectral bands. 

\begin{figure}
    \centering
    \begin{tikzpicture}
    \node[inner sep=0pt] (graph) at (0,-6.5)
        {\includegraphics[width=0.95\linewidth]{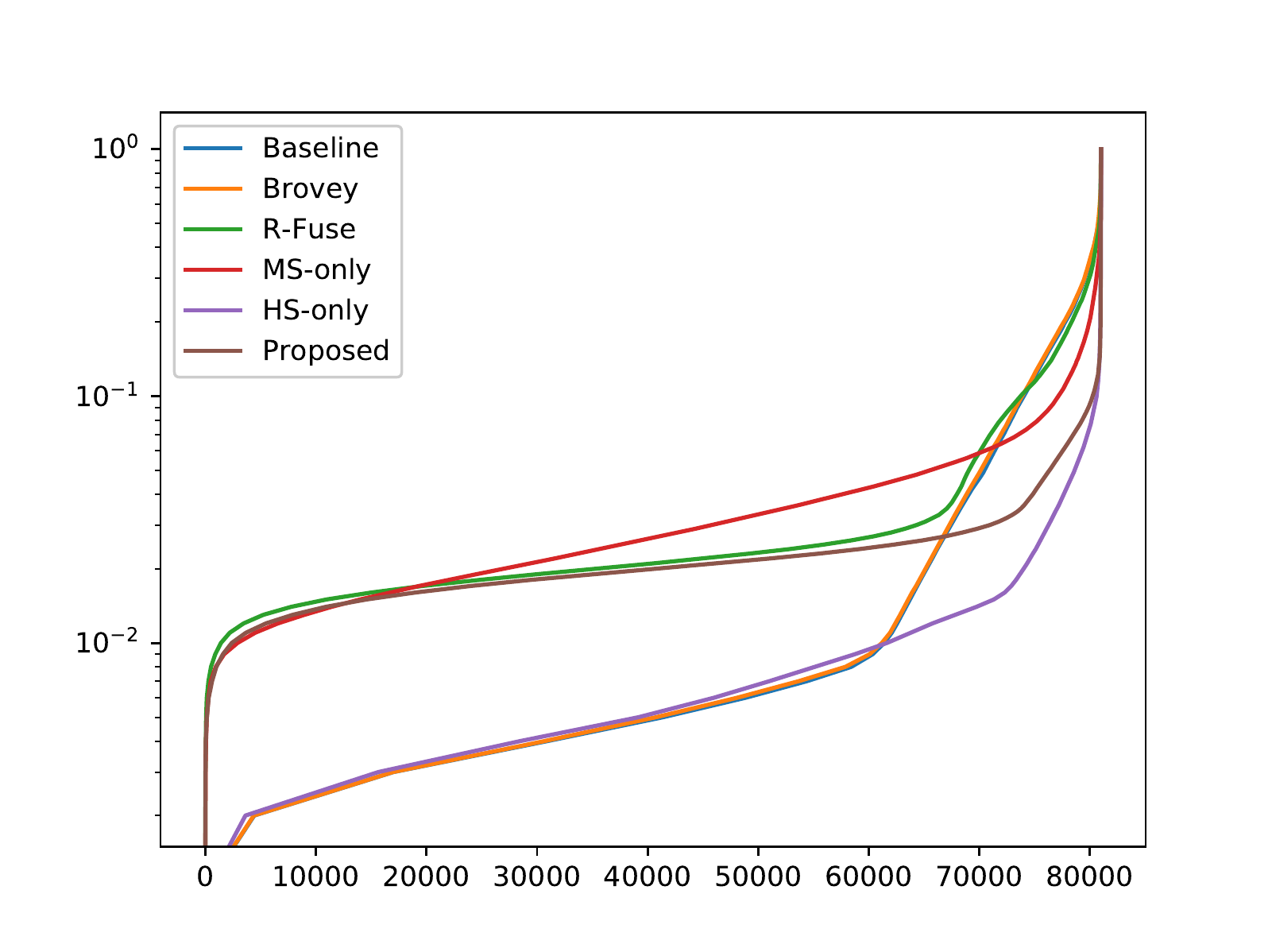}};
    \node[rotate=90] at (-4,-6.5) {\scriptsize SAM};
    \node[] at (0,-9.5) {\scriptsize Number of pixels};

    \end{tikzpicture}
    \caption{SAM cumulative histograms obtained by the compared methods.}
    \label{fig:sam_histcumul}
\end{figure}

\begin{figure}
    \centering
    \begin{tikzpicture}
    \node[inner sep=0pt] (graph) at (0,-6.5)
        {\includegraphics[width=0.95\linewidth]{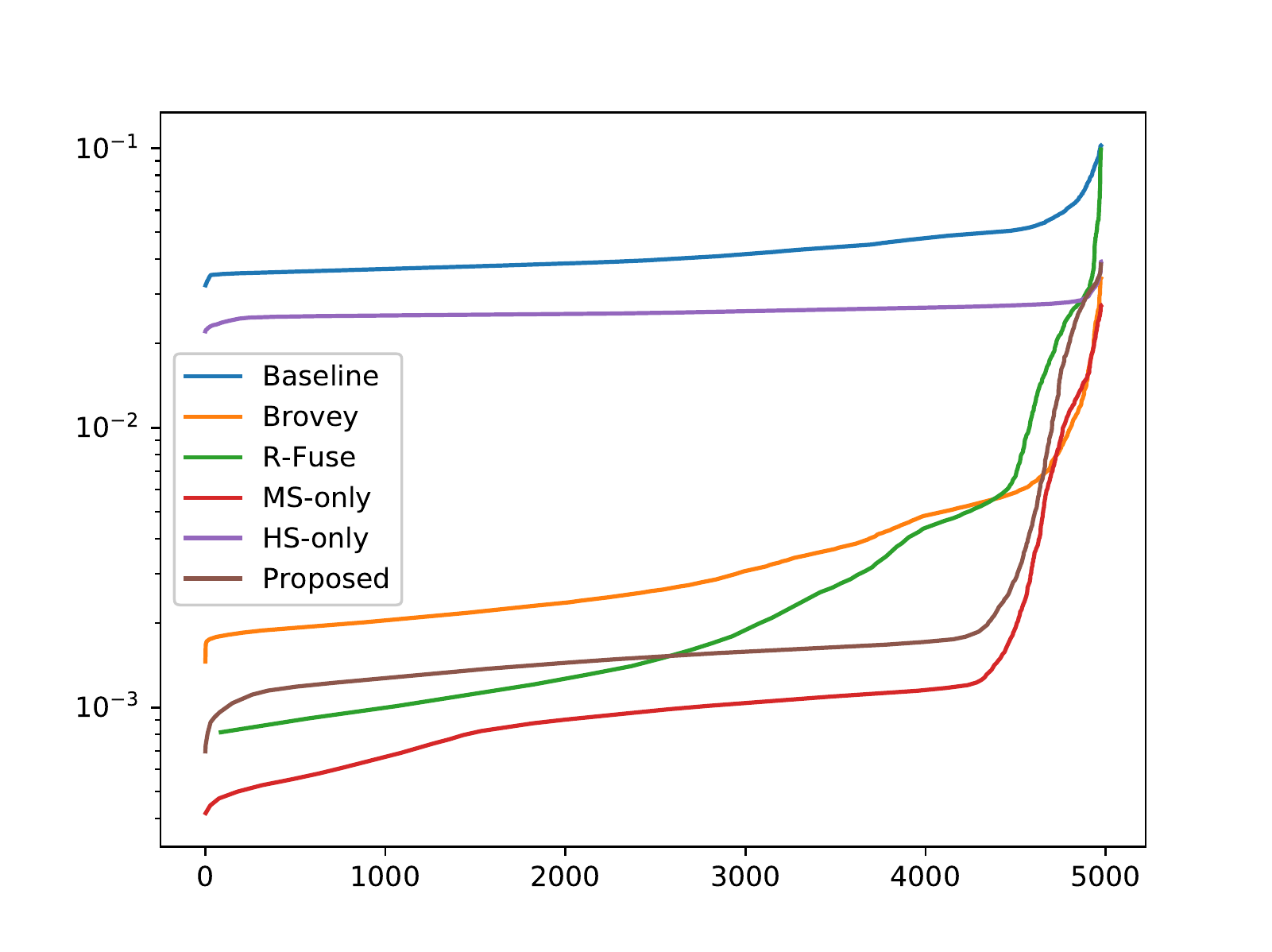}};
    \node[rotate=90] at (-4,-6.5) {\scriptsize cSSIM};
    \node[] at (0,-9.5) {\scriptsize Number of spectral bands};
    
    \end{tikzpicture}
    \caption{cSSIM cumulative histograms obtained by the compared methods.}
    \label{fig:ssim_histcumul}
\end{figure}

Figs.~\ref{fig:sam_histcumul} and \ref{fig:ssim_histcumul} show cumulative histograms of SAM and cSSIM errors respectively. According to Fig.~\ref{fig:sam_histcumul}, R-FUSE appears to provide a high systematic error and a large number of pixels with a SAM value larger than $10^{-1}$rad. On the contrary MS-only and the proposed method also show a high systematic error but a small number of pixels with a large SAM value. On the other hand, the baseline and Brovey present a low systematic error but a large number of pixels with a large SAM value. HS-only shows the best cumulative histogram, with a low systematic error and a small number of pixels with a large SAM value. However in Fig.~\ref{fig:ssim_histcumul}, this method, as well as the baseline, provide a very high (larger than $10^{-2}$) cSSIM value for all spectral bands, while all the other methods show a much lower systematic error. Brovey and MS-only seem to have a very low number of spectral bands with a large cSSIM, but Brovey shows a much larger systematic error. R-FUSE and the proposed method present an intermediate cSSIM cumulative histogram, with a larger number of spectral bands with a high cSSIm value for the R-FUSE method. Considering these two figures, our proposed method emerges once more as a trade-off between good spatial and spectral reconstructions.

\subsection{Selecting the regularization parameter}\label{sect:param}

\begin{figure}
    \centering
    \includegraphics[width=\linewidth]{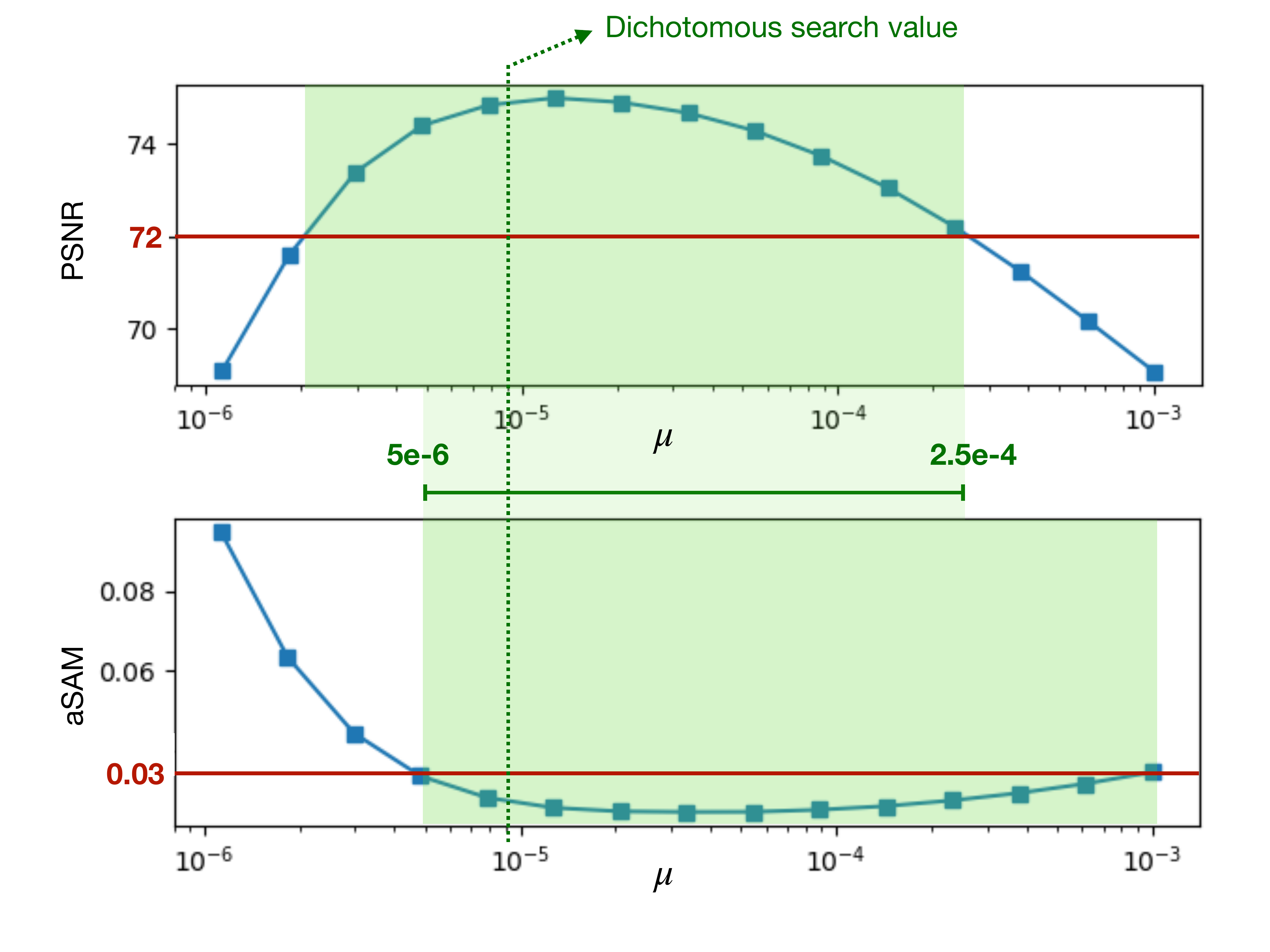}
    \caption{Performance (in terms of PSNR and aSAM) of the proposed fusion algorithm as a function of the regularization parameter $\mu$. Shaded green areas indicate the ranges of values for which the proposed algorithm outperforms state-of-the-art methods. The value of the parameter obtained by the proposed dichotomous search is highlighted with a vertical dotted line.}
    \label{fig:set_mu}
\end{figure}

To choose an appropriate value for the regularization parameter $\mu$ in \eqref{eq:pinv}, we evaluated performances of the proposed fusion algorithm by monitoring the obtained aSAM and PSNR as functions of $\mu$. Results are displayed in Fig. \ref{fig:set_mu}. In the simulations, we selected $\mu = 2.10^{-5}$  as a trade-off between the values providing the best PSNR and aSAM. We see that, for a wide range of $\mu$ values (light green, typically between $5.10^{-6}$ and $2.5.10^{-4}$), the proposed algorithm still outperforms state-of-the-art algorithms. 

In a real-world scenario, i.e., when no ground truth is available and thus quantitative performance measures cannot be computed, we propose to adjust the regularization parameter $\mu$ automatically thanks to a dichotomous approach. More precisely, the optimal value of the parameter is assumed to provide a fused product $\hat{\mathbf{X}}$ such that the residuals defined by the forward models are of magnitude of the noise levels, i.e.,
\begin{eqnarray}
      \|\mathbf{Y}_\textrm{m} -\mathbf{L}_\textrm{m}\mathcal{M}(\hat{\mathbf{X}})\|_\mathrm{F}^2 &\approx& \sigma_\mathrm{m}^2\\
      \|\mathbf{Y}_\textrm{h} -\mathbf{L}_\textrm{h}\mathcal{H}(\hat{\mathbf{X}})\mathbf{S}\|_\mathrm{F}^2 &\approx& \sigma_\mathrm{h}^2.
\end{eqnarray}
Therefore, if the residuals are higher (resp. lower) than the noise levels, we increase (resp. decrease) the value of $\mu$. As illustrated in Fig~\ref{fig:set_mu}, the final value obtained by this iterative procedure is shown to belong to the range of acceptable values.


\section{Discussion and conclusion}
\label{sect:conclusion}
In this paper, we proposed a novel hyperspectral and multispectral image fusion method when the observed images were affected by spectrally variant blurs. To computationally handle this particularity, we elaborated a fast algorithm to minimize the objective function associated with the fusion problem. Operating in the Fourier domain, this algorithm exploited the frequency properties of cyclic convolution operators and capitalized on a low-rank decomposition of the fused image. This implicit spectral regularization allowed the problem to be solved in a subspace of significantly lower dimension. These two computational advantages made the proposed algorithm able to handle large dataset since it solved the fusion problem with reasonable processing times. 
The relevance of the proposed method was evaluated in the specific context of astronomical imaging. We applied this method to a realistic simulated scene of the Orion Bar and compared the results with fused products obtained with state-of-the-art methods and non-symmetric versions of our approach. We showed that the proposed  method appeared as an excellent trade-off with the best spectral, spatial \emph{and} overall reconstruction results. 

Improvements of the fusion method are however required.
Further work will be dedicated to design a tailored regularization term, that could be more suitable than the currently chosen one to our kind of data. In a wider perspective, we also would like to include a realistic noise model in our fusion method. 

\appendices

\section{Vectorized operators}\label{app:vectorization}

To handle the vectorized counterpart $\mathbf{\dot{z}}$ of the DFT of the representation coefficients, the subspace basis matrix $\mathbf{V}$ shoud be rewritten  $\mathbf{\underline{V}} = \mathbf{V} \otimes  \mathbf{I}_{p_\mathrm{m}\times p_\mathrm{m}}$ such that 
$$
\mathbf{\underline{V}}\mathbf{\dot{z}}=\left(
    \begin{array}{c}
         {[ \mathbf{V}\mathbf{\dot{Z}}]}^1 \\
    \vdots \\
         {[ \mathbf{V}\mathbf{\dot{Z}}]}^{l_\mathrm{h}}
    \end{array}
    \right).
$$
Similarly, within this vectorized formulation, the spectral degradation operators ${\mathbf{L}_\mathrm{m}}$ and ${\mathbf{L}_\mathrm{h}}$ should be rewritten as 
\begin{align*}
    \underline{\mathbf{L}_\mathrm{m}} &=\mathbf{L}_\mathrm{m} \otimes  \mathbf{I}_{p_\mathrm{m}} \\
    \underline{\mathbf{L}_\mathrm{h}} &=\mathbf{L}_\mathrm{h} \otimes  \mathbf{I}_{p_\mathrm{h}}.
\end{align*}
Corresponding convolution operators $\underline{\mathbf{\dot{M}}}$ and 
$\underline{\mathbf{\dot{H}}}$ are two block-diagonal matrices
\begin{align*}
    \underline{\mathbf{\dot{M}}} &= \mathrm{diag} \left\{ \mathbf{\dot{M}}^1, \cdots, \mathbf{\dot{M}}^{l_\mathrm{h}}  \right\}\\
    \underline{\mathbf{\dot{H}}} &= \mathrm{diag} \left\{ \mathbf{\dot{H}}^1, \cdots, \mathbf{\dot{H}}^{l_\mathrm{h}}  \right\}
\end{align*}
defined by the DFTs of the MS and HS PSFs along the spectral bands.
Finally, the spatial operators $\underline{\mathbf{\dot{S}}}$ and $\underline{\mathbf{\dot{D}}}$ are written as
\begin{align*}
    \underline{\mathbf{\dot{S}}} &= \mathbf{I}_{l_\mathrm{h}} \otimes \mathbf{\dot{S}}^H \\ 
    \underline{\mathbf{\dot{D}}} &= \mathbf{I}_{l_\mathrm{sub}} \otimes \mathbf{\dot{D}}^H.
\end{align*}

\section{Structure and efficient computation of the linear system matrix $\mathbf{A}$}\label{app:matrix_A}
Capitalizing on the vectorized formulation of the objective function \eqref{eq:pinv2}, the matrix $\mathbf{A}$ defining the linear system to solve exhibits a particular structure. More precisely, as stated by \eqref{eq:amatrix}, $\mathbf{A}$ can be written as a weighted sum of 3 matrices denoted here as $\mathbf{A}_\mathrm{m}$, $\mathbf{A}_\mathrm{h}$ and $\mathbf{A}_\mathrm{r}$ whose computations are discussed in what follows. First, the matrix $\mathbf{A}_\mathrm{m} \triangleq \mathbf{\underline{V}}^H\underline{\mathbf{\dot{M}}}^H\underline{\mathbf{L}}_\mathrm{m}^H\underline{\mathbf{L}}_\mathrm{m}\underline{\mathbf{\dot{M}}}\mathbf{\underline{V}}$ associated with the MS forward model can be decomposed into $l_{\mathrm{sub}} \times l_{\mathrm{sub}}$ elementary blocks such that 
\begin{align*}
    \mathbf{A}_\mathrm{m} 
    &= \left[
    \begin{array}{ccc}
    \left[\mathbf{A}_\mathrm{m}\right]_1^1
    &  \dots & 
    \left[\mathbf{A}_\mathrm{m}\right]^1_
    {l_\mathrm{sub}} \\
    \vdots & \ddots & \vdots \\
    \left[\mathbf{A}_\mathrm{m}\right]^{l_\mathrm{sub}}_1
    &  \dots & 
    \left[\mathbf{A}_\mathrm{m}\right]_{l_\mathrm{sub}}^
    {l_\mathrm{sub}} 
   \end{array} \right]
\end{align*}
where each block $\left[\mathbf{A}_\mathrm{m}\right]_i^j \in \mathbb{R}^{p_\mathrm{m}\times p_\mathrm{m}}$ is a diagonal matrix defined by
\begin{align*}
    \left[\mathbf{A}_\mathrm{m}\right]_i^j  = \mathrm{diag} \left\{ \sum_{l=1}^{l_\mathrm{m}} \boldsymbol{\alpha}_i^l
    \odot
    \bar{\boldsymbol{\alpha}}_j^l \right\} 
\end{align*}
with $\boldsymbol{\alpha}_j^l = \sum_{b=1}^{l_\mathrm{h}} [\mathbf{L}_\mathrm{m}]_b^l \mathbf{\dot{M}}^b \mathbf{V}_j^b$. This computation is detailed in Algo. \ref{algo:am}. Since the matrix $\mathbf{A}_\mathrm{m}$ is symmetric, note that only its upper (or lower) triangular part needs to be calculated.

\begin{algorithm}[h!]
\caption{Computing $\mathbf{A}_\mathrm{m}$ }
 \begin{algorithmic}[1]
 \renewcommand{\algorithmicrequire}{\textbf{Input:} }
 \renewcommand{\algorithmicensure}{\textbf{Output:} }
 \REQUIRE $\mathbf{L}_\mathrm{m}$, $\mathbf{V}$, $\mathbf{\dot{M}}$
     \\ \textit{\# Compute all $\boldsymbol{\alpha}_j^l$ ($\forall j,l$)}
  \FOR {$l = 1$ to $l_\mathrm{m}$}
    \FOR {$j = 1$ to $l_\mathrm{sub}$}
        \STATE $\boldsymbol{\alpha}_j^l$ = $\sum_{b=1}^{l_\mathrm{h}} [\mathbf{L}_\mathrm{m}]_b^l \mathbf{\dot{M}}^b \mathbf{V}_j^b$
    \ENDFOR
  \ENDFOR
  \\ \textit{\# Fill-in $\mathbf{A}_\mathrm{m}$ block-by-block}
  \FOR {$i = 1$ to $l_\mathrm{sub}$}
    \FOR {$j = 1$ to $\left( l_\mathrm{sub}-i \right)$}
          \IF {($j \ne 0$)}
          \STATE $[\mathbf{A}_\mathrm{m}]_i^{j+i} = \sum_{l=1}^{l_\mathrm{m}} \boldsymbol{\alpha}_i^l \odot \bar{\boldsymbol{\alpha}}_j^l$
          \vspace{0.1cm}
          \STATE $[\mathbf{A}_\mathrm{m}]_{j+i}^i = \overline{[\mathbf{A}_\mathrm{m}]_i^{j+i}}$
          \ELSE
          \STATE $[\mathbf{A}_\mathrm{m}]_i^{i} = \sum_{l=1}^{l_\mathrm{m}} \boldsymbol{\alpha}_i^l \odot \bar{\boldsymbol{\alpha}}_i^l$
          \ENDIF
    \ENDFOR
  \ENDFOR
 \ENSURE  $\mathbf{A}_\mathrm{m}$
 \end{algorithmic} 
 \label{algo:am}
 \end{algorithm}
 
Similarly, the matrix $\mathbf{A}_\mathrm{h} \triangleq \mathbf{\underline{V}}^H\underline{\mathbf{\dot{H}}}^H\underline{\mathbf{L}}_\mathrm{h}^H\underline{\mathbf{\dot{S}}}^H \underline{\mathbf{\dot{S}}}\underline{\mathbf{L}}_\mathrm{h}\underline{\mathbf{\dot{H}}}\mathbf{\underline{V}}$ defined by the HS forward model can be decomposed into $l_{\mathrm{sub}} \times l_{\mathrm{sub}}$ elementary blocks such that
\begin{align*}
\mathbf{A}_\mathrm{h}  
    = \left[
    \begin{array}{ccc}
    \left[\mathbf{A}_\mathrm{h}\right]_1^1
    &  \dots & 
    \left[\mathbf{A}_\mathrm{h}\right]^1_
    {l_\mathrm{sub}} \\
    \vdots & \ddots & \vdots \\
    \left[\mathbf{A}_\mathrm{h}\right]^{l_\mathrm{sub}}_1
    &  \dots & 
    \left[\mathbf{A}_\mathrm{h}\right]_{l_\mathrm{sub}}^
    {l_\mathrm{sub}} 
   \end{array} \right]
\end{align*}
where each block $\left[\mathbf{A}_\mathrm{h}\right]_i^j \in \mathbb{R}^{p_\mathrm{m}\times p_\mathrm{m}}$ is also decomposed into $d^2 \times d^2$ diagonal matrices of size $\sfrac{p_\mathrm{m}}{d^2}\times \sfrac{p_\mathrm{m}}{d^2}$, i.e., 
\begin{align*}
\left[\mathbf{A}_\mathrm{h}\right]_i^j =
\boldsymbol{\beta}_{i}^{j} \odot \boldsymbol{\Upsilon}_{p_\mathrm{m},{d^2}}
\end{align*}
with $\boldsymbol{\beta}_{i}^{j} = \frac{1}{d^2} \sum_{l=1}^{l_\mathrm{h}} \left([\mathbf{L}_\mathrm{h}]_l^l\right)^2 \mathbf{V}_j^l \mathbf{V}_i^l (\mathbf{\dot{H}}^l) (\mathbf{\dot{H}}^l)^H$ and
\begin{equation}
 \boldsymbol{\Upsilon}_{p_\mathrm{m},{d^2}}    = \left[\begin{array}{ccc}  \mathbf{I}_{\sfrac{p_\mathrm{m}}{d^2}} & \dots & \mathbf{I}_{\sfrac{p_\mathrm{m}}{d^2}} \\
\vdots & \ddots & \vdots \\ \mathbf{I}_{\sfrac{p_\mathrm{m}}{d^2}}
&  \dots & 
\mathbf{I}_{\sfrac{p_\mathrm{m}}{d^2}}
\end{array} \right] 
\end{equation}
Note that a large number of coefficients in $\mathbf{A}_\mathrm{h}$ are zeros, which avoids to compute all the entries in the matrices $\boldsymbol{\beta}_{i}^{j}$ but only its non-zero coefficients whose positions correspond to the non-zero values in $\boldsymbol{\Upsilon}_{p_\mathrm{m},{d^2}}$. This is summarized in Algo. \ref{algo:ah} which also benefits from the Hermitian symmetry of $\mathbf{A}_\mathrm{h}$.

\begin{algorithm}[h!]
 \caption{Computing $\mathbf{A}_\mathrm{h}$}
 \begin{algorithmic}[1]
 \renewcommand{\algorithmicrequire}{\textbf{Input:} }
 \renewcommand{\algorithmicensure}{\textbf{Output:} }
 \REQUIRE $\mathbf{L}_\mathrm{h}$, $\mathbf{V}$, $\mathbf{\dot{H}}$, $d$
  \\ \textit{\# Fill-in $\mathbf{A}_\mathrm{h}$ block-by-block}
  \FOR {$i = 1$ to $l_\mathrm{sub}$}
    \FOR {$j = 1$ to $\left( l_\mathrm{sub}-i \right)$}
          \IF {($j \ne 0$)}
           \STATE \textit{\# Identify non-zero elements in the block $(_i^j)$}
          \FOR {$(m,n)$ s.t. $[\boldsymbol{\Upsilon}_{p_{\mathrm{m}},d^2}]_m^n$ $\ne 0$}
             \STATE $\left[\mathbf{A}_\mathrm{h}\right]_{i,m}^{i+j,n} = 
             \frac{1}{d^2} \sum_{l=1}^{l_\mathrm{h}} \left([\mathbf{L}_\mathrm{h}]_l^l\right)^2 \mathbf{V}_j^l \mathbf{V}_i^l \mathbf{\dot{H}}_m^l \mathbf{\dot{H}}_n^l$
             \vspace{0.1cm}
             \STATE $\left[\mathbf{A}_\mathrm{h}\right]_{i+j}^i = \overline{\left(\mathbf{A}_\mathrm{h}\right)_{i}^{i+j}}$
          \ENDFOR
          \ELSE
        \STATE \textit{\# Identify non-zero elements in the block $(_i^i)$}
          \FOR {$(m,n)$ s.t. $[\boldsymbol{\Upsilon}_{p_{\mathrm{m}},d^2}]_m^n$ $\ne 0$}
            \STATE $\left[\mathbf{A}_\mathrm{h}\right]_{i,m}^{i,n} = \frac{1}{d^2} \sum_{l=1}^{l_\mathrm{h}} \left([\mathbf{L}_\mathrm{h}]_l^l\right)^2 \mathbf{V}_i^l \mathbf{V}_i^l \mathbf{\dot{H}}_m^l \mathbf{\dot{H}}_n^l$
          \ENDFOR
          \ENDIF
    \ENDFOR
  \ENDFOR
 \ENSURE  $\mathbf{A}_\mathrm{h}$
 \end{algorithmic} 
 \label{algo:ah}
 \end{algorithm}

Finally, the last matrix involved in the definition of $\mathbf{A}$ is $\mathbf{A_\mathrm{r}} \triangleq \underline{\mathbf{\dot{D}}}^H\underline{\mathbf{\dot{D}}}$, which is a diagonal matrix and easily computable. 


\bibliographystyle{IEEEtran}
\bibliography{strings_all_ref,refs}

\end{document}